\documentclass{article}

\usepackage{PRIMEarxiv}

\usepackage[utf8]{inputenc} 
\usepackage[T1]{fontenc}    
\usepackage{hyperref}       
\usepackage{url}            
\usepackage{booktabs}       
\usepackage{amsfonts}       
\usepackage{nicefrac}       
\usepackage{microtype}      
\usepackage{lipsum}
\usepackage{enumitem}
\usepackage{float}
\usepackage[misc]{ifsym}    
\usepackage{graphicx}
\usepackage{multirow} 
\usepackage{makecell}
\usepackage{amsmath, amssymb}
\usepackage{hhline} 
\usepackage{adjustbox} 
\graphicspath{{media/}}     
\usepackage{pdfpages} 
\usepackage{longtable} 

\usepackage{authblk}
\usepackage[backend=biber, sorting=none, defernumbers=false]{biblatex}
\title{MoMA: A Mixture-of-Multimodal-Agents Architecture for Enhancing Clinical Prediction Modelling}

\author[1]{Jifan Gao}
\author[1]{Mahmudur Rahman}
\author[1]{John Caskey}
\author[1]{Madeline Oguss}
\author[1]{Ann O'Rourke}
\author[1]{Randy Brown}
\author[2]{Anne Stey}
\author[1]{Anoop Mayampurath}
\author[1]{Matthew M. Churpek}
\author[1,*]{Guanhua Chen}
\author[1,*]{Majid Afshar}

\affil[1]{University of Wisconsin-Madison, Madison, Wisconsin, USA}
\affil[2]{Northwestern University, Chicago, Illinois, USA}
\affil[*]{Corresponding authors: gchen25@wisc.edu, mafshar@medicine.wisc.edu}

\begin{document}
\maketitle

\begin{abstract}
Multimodal electronic health record (EHR) data provide richer, complementary insights into patient health compared to single-modality data. However, effectively integrating diverse data modalities for clinical prediction modeling remains challenging due to the substantial data requirements. We introduce a novel architecture, Mixture-of-Multimodal-Agents (MoMA), designed to leverage multiple large language model (LLM) agents for clinical prediction tasks using multimodal EHR data. MoMA employs specialized LLM agents (``specialist agents'') to convert non-textual modalities, such as medical images and laboratory results, into structured textual summaries. These summaries, together with clinical notes, are combined by another LLM (``aggregator agent'') to generate a unified multimodal summary, which is then used by a third LLM (``predictor agent'') to produce clinical predictions. Evaluating MoMA on three prediction tasks using real-world datasets with different modality combinations and prediction settings, MoMA outperforms current state-of-the-art methods, highlighting its enhanced accuracy and flexibility across various tasks.
\end{abstract}

\keywords{Multimodal learning \and Large language models \and Multi-agent system \and Clinical predictions}

\begin{refsegment}

\section*{Introduction}
\label{sec:introduction}

Modern healthcare increasingly leverages electronic health records (EHRs), which integrate diverse patient data modalities, such as clinical notes, medical images, vital signs, and laboratory results \cite{cai2019survey}. Each modality contributes unique, complementary information: clinical notes summarize patient symptoms, diagnoses, and treatments as documented by healthcare professionals; medical images objectively depict anatomical structures and pathology, facilitating disease detection and monitoring; laboratory and vital sign data quantify physiological states and abnormalities. The integration of multimodal EHR data enables a more holistic understanding of a patient’s health conditions. The adaptation of multimodal EHR into a machine learning pipeline has been demonstrated to outperform those that only leverage a single modality in a wide range of clinical prediction tasks \cite{rohaut2024multimodal, soenksen2022integrated, winston2024multimodal, gao2024automated, kline2022multimodal}.


Multimodal integration methodologies typically fall into three categories: early fusion (concatenating inputs before training), joint fusion (co-learning representations during training), and late fusion (combining outputs from separately trained models) \cite{acosta2022multimodal, huang2020fusion}. Joint fusion is particularly promising, as it facilitates co-learning of a shared vector space, effectively capturing intricate cross-modal relationships. Various methods have been developed for learning this shared vector space, including cross-attention mechanisms \cite{li2022blip}, mixture-of-expert framework \cite{han2024fusemoe}, contrastive learning \cite{radford2021learning}, masked vision/language modeling \cite{bannur2023learning}, and variational approaches \cite{cohen2023joint}. This method has consistently demonstrated superior performance compared to early and late fusion strategies across diverse clinical applications \cite{guarrasi2024systematic, stahlschmidt2022multimodal, hayat2022medfuse}. 

In particular, the recent surge of multimodal large language models (LLMs) is advancing the development of joint fusion methods. Vision language models, such as BLIP-2 \cite{li2023blip}, Flamingo \cite{alayrac2022flamingo}, Kosmos-2 \cite{peng2023kosmos}, and PaLM-E \cite{driess2023palm}, have achieved state-of-the-art results while learning unified representations for images (or videos) and text from paired corpora. Beyond vision and text, ImageBind \cite{girdhar2023imagebind} and OneLLM \cite{han2024onellm} expand the shared space to six or more modalities, delivering competitive performance by contrastively aligning each additional signal to an image- or language-anchored representation. In the medical domain, LLaVA-Med \cite{li2024llava}, VILA-M3 \cite{nath2025vila}, and GSCo \cite{he2024gsco} align radiology images with paired reports and achieved strong performance in various tasks. 

All of these approaches, however, still depend on large paired multimodal datasets to learn a joint vector space, requiring non-trivial supervised alignment when a new modality is introduced incrementally. In healthcare, obtaining sufficient paired data is challenging due to the complexities associated with linking distinct modality-specific resources to the same patient or clinical encounter \cite{boehm2022harnessing}, coupled with inherent data fragmentation in healthcare systems \cite{chang2025continuous}. These limitations present substantial obstacles to developing accurate multimodal models when lacking sufficient paired data for learning the shared vector space across various modalities.

Given these limitations, the fundamental motivation behind our method is to leverage the inherent capability of pretrained LLMs to translate multimodal clinical data into natural language. LLMs are widely recognized for their capability to capture semantic meaning from text \cite{zhao2023survey, wei2022emergent}, including clinical notes \cite{thirunavukarasu2023large}. When provided with clinical text inputs, LLMs can also serve as classifiers in clinical prediction tasks and have achieved outstanding performance \cite{wang2024drg, liu2023medical, gu2024probabilistic}. Modern multimodal LLMs can also understand modalities beyond plain text, including medical images \cite{li2024llava, lee2023cxr} and structured EHR data \cite{zhu2024prompting, gao2024raw}, and can convert non-plain text data into text summaries. Because of these rapid advances in multimodal LLMs, which demonstrate that rich semantics from non-text sources can be effectively translated into natural language, and the view from cognitive theory that language is central to human cognition \cite{lupyan2016centrality}, the converted text can serve as an aligned space, analogous to the shared vector space used in traditional joint-fusion multimodal approaches. Importantly, this conversion can be performed in a zero-shot manner using pretrained LLMs, avoiding the extensive data requirements typically associated with constructing such vector spaces.

\begin{figure}[ht!]
    \centering
    \includegraphics[width=\textwidth]{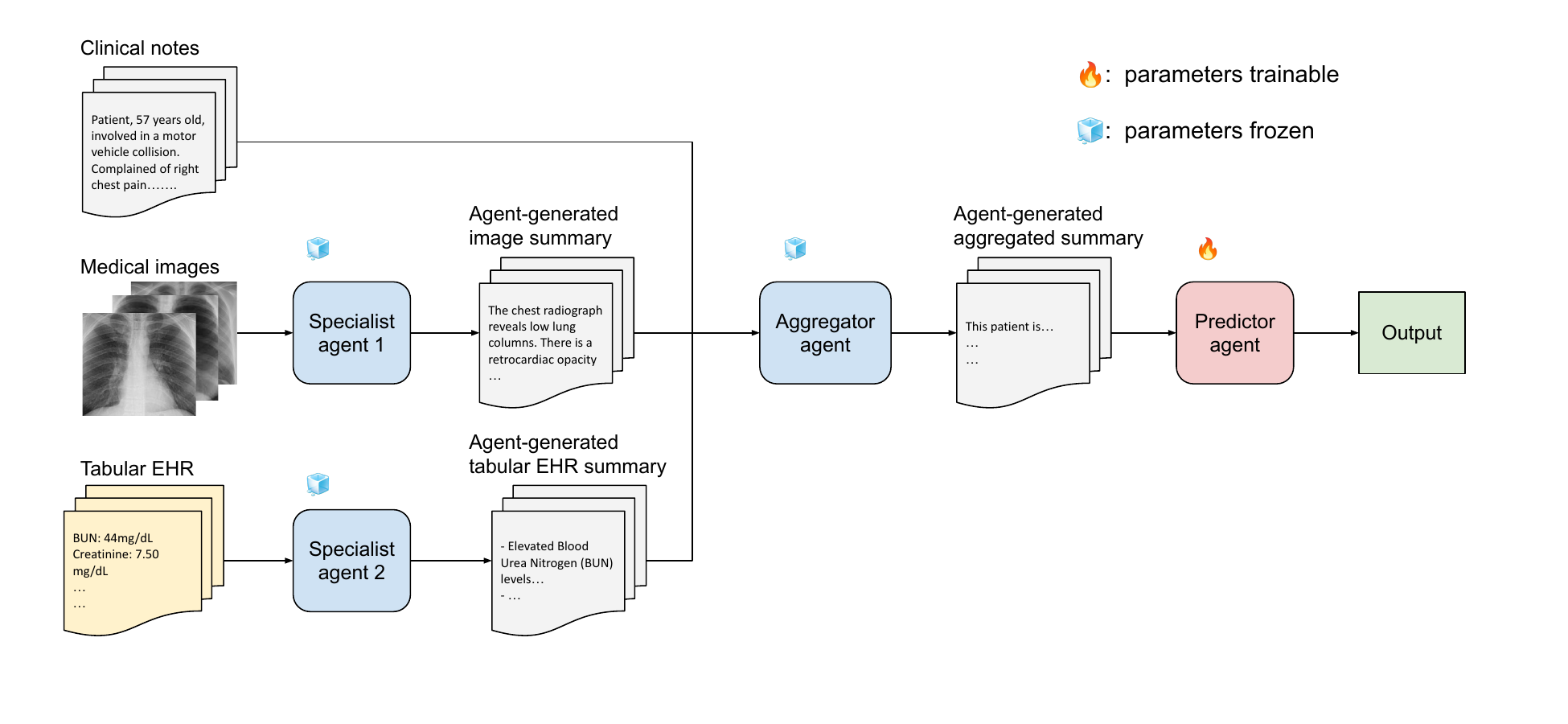}
    \caption{Architecture of MoMA. Non-plain-text modalities, including medical images and tabular EHR data, are transformed into text summaries by specialist agents. These agent-generated summaries are then appended to the original clinical notes and passed to the aggregator agent, which produces a comprehensive and concise summary from the combined text. This aggregated summary is subsequently fed into a predictor agent to generate output predictions. Given the well-known capabilities of LLMs in clinical summarization, both the specialist and aggregator agents operate in a zero-shot setting, with only the predictor agent requiring training or fine-tuning, which reduces computational costs during the training phase.}
    \label{fig:architecture}
\end{figure}

In addition, studies have introduced the collaborative potential of LLMs whereby an LLM agent produces an improved response when incorporating outputs from other LLM agents \cite{wang2024mixture}. This discovery suggests that introducing additional LLM agents, following the multimodal LLMs, to integrate original clinical text with LLM-generated summaries from non-text modalities could benefit predictive performance. Motivated by this observation, we propose the Mixture-of-Multimodal-Agents (MoMA) architecture for clinical prediction with multimodal EHR data (Figure \ref{fig:architecture}). In MoMA, each non-text modality is processed by a pretrained, modality-specialized LLM agent that converts the non-text data into a corresponding text summary. These generated text summaries, along with existing clinical notes, are then integrated by an aggregator agent to form a unified narrative, which is used by a final predictor agent to generate clinical predictions. Details of the MoMA architecture are described in the \nameref{sec:methods} section. 

The sequential transfer of these summaries harnesses the collaborative potential of LLMs, facilitating an effective integration of multimodal EHR data. MoMA can immediately leverage advances in any multimodal LLMs without retraining, because the non-text to text conversion can be swapped without retraining the rest of the system. Unlike existing multimodal LLMs that require paired multimodal data to learn a shared vector space, MoMA is an architecture that incorporates existing state-of-the-art multimodal LLMs in a plug-and-play fashion. Moreover, this architecture reduces training requirements by allowing the specialist and aggregator agents to operate in a zero-shot manner, with only the predictor agent requiring fine-tuning.

We validated the MoMA architecture on real-world, private datasets across three clinical tasks (chest trauma severity stratification, multitask chest and spine trauma severity stratification, and unhealthy alcohol use screening) with different combinations of modalities. Our results show that MoMA not only achieves superior performance in the overall testing set but also outperforms baseline models in every sex and race subgroup. Our ablation study reveals that the performance improvements benefit not only from the LLMs' text understanding capabilities but also from the effective integration of non-text modalities.

Traditional approaches to developing multimodal prediction models in healthcare require large volumes of high-quality paired EHR data to learn effective joint fusion representations, a requirement that is often unmet due to data quality challenges and strict privacy regulations that complicate the sharing of pretraining models \cite{li2023ethics, haltaufderheide2024ethics}. By leveraging open-source LLMs, our MoMA architecture circumvents these obstacles by translating non-text modalities into the natural language space without the need for extensive paired datasets. This not only reduces the resource burden associated with traditional joint fusion methods but also enables institutions with limited access to comprehensive multimodal data to develop accurate clinical prediction models. 

\section*{Results}


\subsection*{Datasets and Cohort Characteristics}
\label{subsec:datasets}
We validated the MoMA architecture on three clinical prediction tasks using private datasets collected from the University of Wisconsin Hospitals and Clinics (UW Health): chest trauma severity stratification, multitask chest and spine trauma severity stratification, and unhealthy alcohol use screening. These tasks differ in complexity and classification structure: the first involves multiclass classification for chest trauma, the second jointly predicts multiclass severity for both chest and spine, and the third addresses binary classification for unhealthy alcohol use. They also involve distinct modality combinations: the first two tasks integrate free text clinical notes and chest radiographs, while the unhealthy alcohol use screening task combines free text and lab measurements.

Traumatic injuries are the leading cause of death among people younger than 45 \cite{herrera2022survival}. Chest trauma is one of the most commonly encountered trauma injuries. Nearly half of trauma-related deaths occur after hospital admission \cite{lefering2012epidemiology} and timely stratification of chest trauma injury severity can help triage patients and predict complications \cite{granstrom2018criteria}. The cohort used for the chest trauma severity stratification task was collected between January 2015 and December 2019, with a total sample size of 2,722 unique patients. This task involved a three-class classification of injury severity, annotated as negative, minor/moderate, and serious or greater. A team of certified trauma registrar coders at UW Health conducted extensive manual chart abstraction for each patient encounter, adhering to American College of Surgeons (ACS) and Trauma Quality Improvement Program (TQIP) standards \cite{shafi2009trauma} to calculate and validate Abbreviated Injury Scale (AIS) scores and associated trauma metrics \cite{palmer2016defining}. Each patient encounter includes clinical notes as the text modality and chest radiographs as the non-text modality from the EHR.

Using the same cohort from the chest trauma severity stratification task, we defined a more complex multitask setting where models simultaneously predict injury severity for both the chest and spine. Each encounter is labeled using the same annotation protocol as the chest trauma severity stratification task, with injury severity assessed for the chest and spine. Each sample contains clinical notes as the text modality and chest radiographs as the non-text modality. 

We also validated MoMA in an unhealthy alcohol use screening task. Alcohol misuse is recognized by the World Health Organization as one of the top five risk factors contributing to disease burden \cite{jinyi2024global} and timely screening for unhealthy alcohol use can help mitigate the risk of alcohol-related harm \cite{coulton2011alcohol}. In a prospective study, two research teams were deployed to screen, consent, and enroll 2,096 consented patients between September 2021 and February 2024, into the Tobacco, Alcohol, Prescription Medications, and other Substance  (TAPS) screening tool, recommended by the National Institute on Drug Abuse \cite{mcneely2016performance}, to assess unhealthy alcohol use in the past three months. This task uses whether a patient had unhealthy alcohol use in the previous three months as labels for binary classification. The emergency department (ED) recruitment team, comprising trained coordinators, screened willing patients for eligibility. Upon admission, an addiction medicine research team approached eligible patients to obtain informed consent and administer the TAPS screening tool, with participants receiving a gift card upon completion. The manual screen results were collected in a survey database and linked to the patient's related EHR encounter containing clinical notes as the text modality and lab measurements in tabular format as the non-text modality from the EHR.

We conducted temporal validation on both tasks to ensure the test sets are independent of the development set. For chest trauma severity stratification and multitask chest and spine trauma severity stratification, the development set used data collected from January 2015 to December 2018 and the test set used data collected from January 2019 to December 2019. For unhealthy alcohol use screening, the development set used data collected from September 2021 to August 2023 and the test set used data collected from September 2023 to January 2024. The characteristics of the two cohorts are shown in Table \ref{tab:cohort_char}.

\subsection*{Overall Performance}
\label{sebsec:overall performance}

For the chest trauma severity stratification task and the multitask chest and spine trauma stratification, we fine-tuned ClinicalBERT \cite{alsentzer2019publicly} with free text, following the methodology described in Gao et al. \cite{gao2024automated} as a published baseline. For unhealthy alcohol use screens, we compared our methods to a trained 1-dimensional convolutional neural network (1D-CNN) model from Afshar et al. \cite{afshar2022development}, which processes clinical text mapped to medical concepts from the National Library of Medicine (CUI; concept unique identifier). These published baselines represent the current state-of-the-art (SOTA) approach for the three tasks (see the \nameref{sec:methods} section for more detailed descriptions). We also compared MoMA to LLaVA-Med \cite{li2024llava}, a widely used and representative multimodal LLM baseline in the medical domain \cite{guo2024prompting,zhu2025guiding,yang2025medical}, on both the chest trauma severity stratification task and the multitask chest and spine trauma stratification task. LLaVA-Med models were fine-tuned using the development set, with details described in the \nameref{sec:methods} section. Note that LLaVA-Med is not applicable to the unhealthy alcohol use screening task, as it does not support tabular data inputs. In addition, we evaluated the MoMA approach against two vector-based multimodal fusion methods: one using a cross-attention module to integrate representation vectors from two different modalities\cite{jian2024rethinking}, and the other using a Mixture-of-Experts (MoE) mechanism for multimodal fusion\cite{zheng2024multimodal}. Details of these two approaches are described in the \nameref{sec:methods} section. For the cross-attention and MoE baselines of both tasks and the published baseline of the chest trauma severity stratification and the multitask chest and spine trauma severity stratification task, we used the development and test sets as described in Table \ref{tab:cohort_char} to reimplement these methods. While we had access to the trained 1D-CNN model for unhealthy alcohol use screen positives, we did not have access to the original training dataset due to data restrictions. Therefore, the published baseline model had the advantage of being trained on a much larger dataset with a total of 54,915 encounters. 

Since macro- and micro-F1 scores are standard metrics for multiclass classification \cite{gao2024clinical, ouyang2019analysis}, we used them to evaluate chest trauma severity stratification and multitask chest and spine trauma stratification. For the binary task of unhealthy alcohol use screening, we used AUROC (Area Under the Receiver Operating Characteristic Curve) and AUPR (Area Under the Precision-Recall Curve), which are widely used in binary classification tasks \cite{bergquist2023evaluation, bergquist2023framework}. 

\begin{figure}[ht!]
    \centering
    \includegraphics[width=\textwidth]{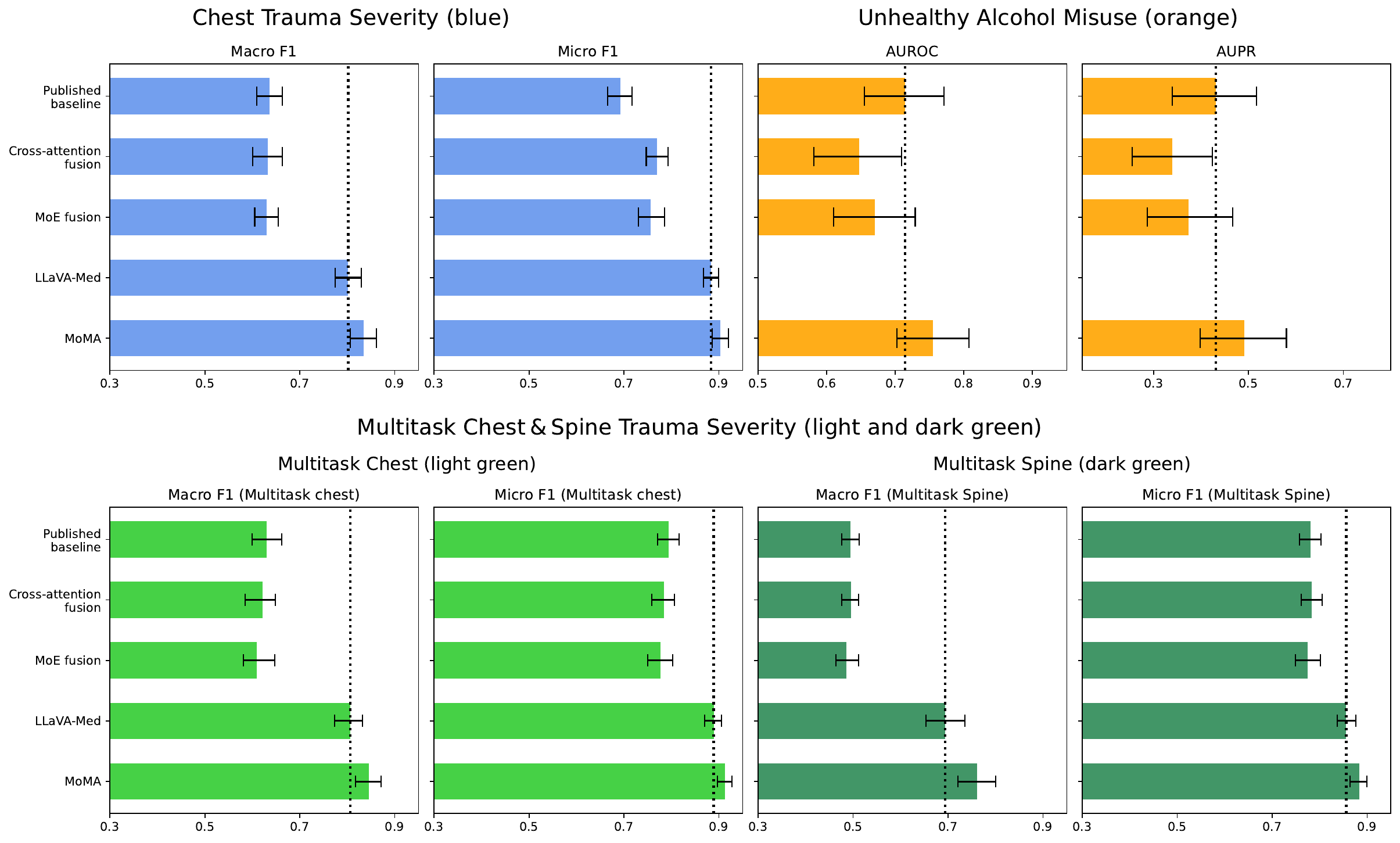}
    \caption{Comparison of discriminative performance and 95\% confidence intervals between MoMA, the published SOTA baseline, and the cross-attention, MoE-based fusion, and fine-tuned LLaVA-Med. Dotted lines mark the strongest baselines, LLaVA-Med for chest trauma and multitask chest and spine trauma stratification, and the published baseline for the unhealthy alcohol use screening.  Confidence intervals were calculated using bootstrap methods. Note that LLaVA-Med accepts only text + image inputs, so it cannot be evaluated on the unhealthy alcohol use screening task, which combines text with tabular data. MoMA consistently outperforms all applicable baselines across these clinical tasks.}
    \label{fig:overall_performance}
\end{figure}

The performance metrics are summarized in Figure \ref{fig:overall_performance}, where best baseline results, LLaVA-Med for chest trauma and multitask trauma stratification, and the published baseline for the unhealthy alcohol use screening, are highlighted using dotted lines for direct comparison; corresponding numerical results are provided in \ref{appendix:main_results}. On both the chest trauma and multitask trauma stratification tasks, the MoMA outperformed all baselines, including the fine-tuned LLaVA-Med. Notably, it greatly outperformed the published baselines as well as the cross-attention and MoE baselines. In particular, for chest trauma severity stratification, both as a single task and within the multitask setting, MoMA achieved macro-F1 scores near 0.85 and micro-F1 scores above 0.90. For spine trauma stratification in the multitask setting, it achieved macro-F1 scores above 0.75 and micro-F1 scores close to 0.90. While a third modality (lab measurement) to MoMA was feasible for the trauma severity stratification tasks, we found that performance had already saturated with two modalities, and adding tabular data with laboratory results offered no additional gains (see \ref{tab:third_modality}). For the unhealthy alcohol use screen task, where LLaVA-Med is not applicable, the published baseline performed better than the cross-attention and MoE methods, as expected given its training on a larger dataset. We also retrained a 1D-CNN model, following the approaches in the published baseline, on the same small cohort used by MoMA and observed that it underperformed (AUROC 0.641, AUPR 0.325, see \ref{tab: 1d_cnn-sample_size}) relative to the published baseline model trained on the larger dataset. Despite these, MoMA achieved stronger results, with an AUROC of around 0.75 and an AUPR near 0.50.

To provide a comprehensive evaluation, we also reported macro-AUROC for the chest trauma severity and multitask chest and spine trauma stratification tasks, as well as F1 scores for the unhealthy alcohol use screening task in the \ref{appendix:main_results}. These additional results are consistent with the trends shown in Figure \ref{fig:overall_performance}.

\subsection*{Subgroup Analysis}

\begin{figure}[ht!]
    \centering
    \includegraphics[width=\textwidth]{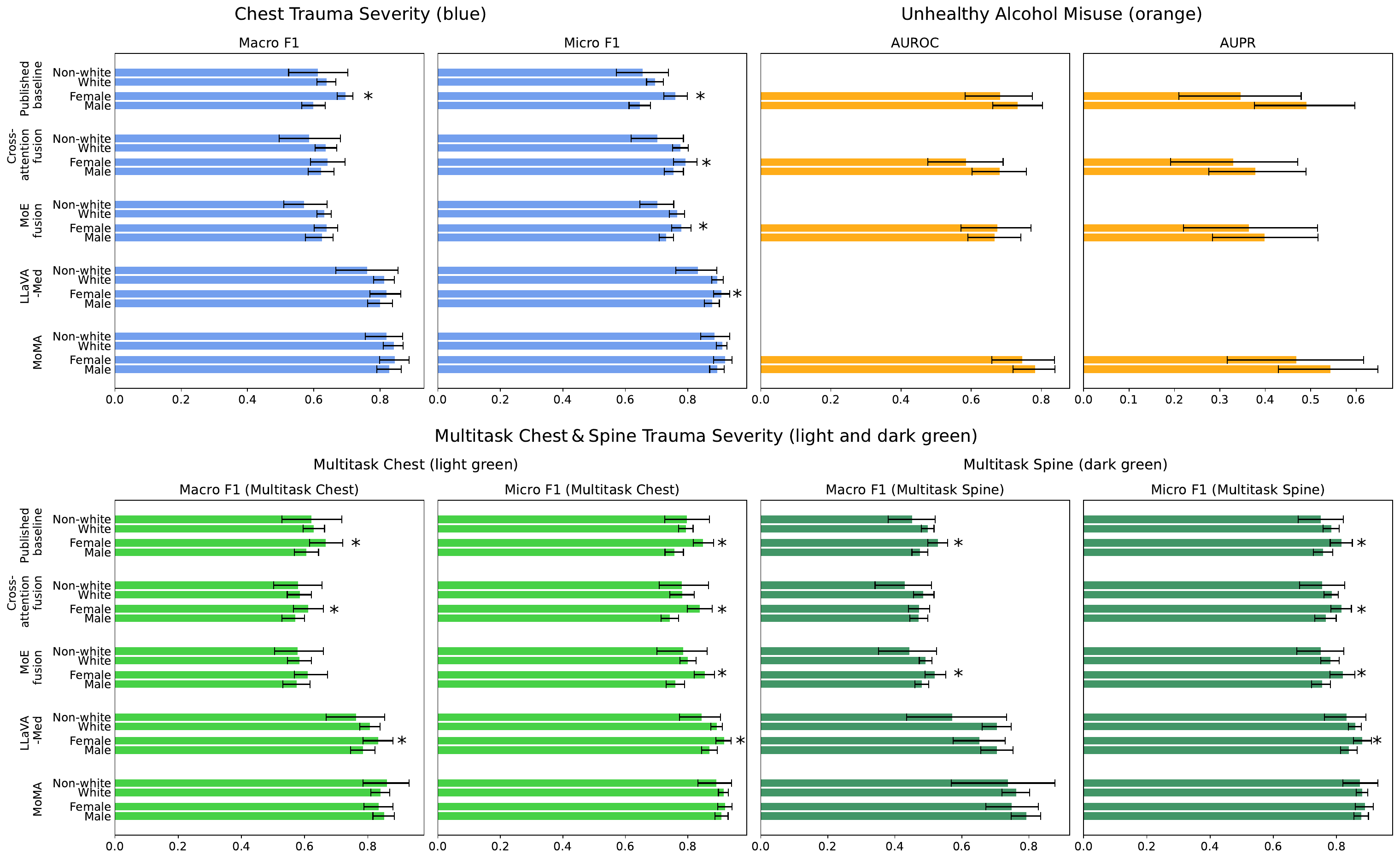}
    \caption{Comparison of model performance across racial and sex subgroups. We excluded the comparison for the white vs. non-white subgroups in the unhealthy alcohol use screening task due to the limited number of cases in the non-white subgroup. Statistically significant differences in performance within each group are marked with asterisks. MoMA achieves the highest and consistent performance across all subgroups in these clinical tasks.}
    \label{fig:subgroup}
\end{figure}

Subgroup analysis is essential in clinical prediction models to ensure consistent performance across patient populations. By evaluating model performance across subpopulations, such as different racial and sex groups, researchers can identify variations in performance that may affect certain populations. The subgroup analysis results are presented in Figure \ref{fig:subgroup} (see also \ref{tab: trauma_subgroup} and \ref{tab: alcohol_subgroup}). Because the limited number of non-white cases in the unhealthy alcohol use screening cohort led to high variance, we excluded the comparison for the white vs. non-white subgroups. Across all other subgroups, the MoMA architecture consistently achieved the best performance. Furthermore, we conducted paired t-tests to compare performance between females and males and between non-white and white individuals. MoMA demonstrated consistent performance across subgroups, while the baseline methods showed notable differences in subgroup performance in the chest trauma severity stratification and the multitask chest and spine trauma severity stratification tasks.

\subsection*{Ablation Study}

\begin{figure}[ht!]
    \centering
    \includegraphics[width=0.90\textwidth]{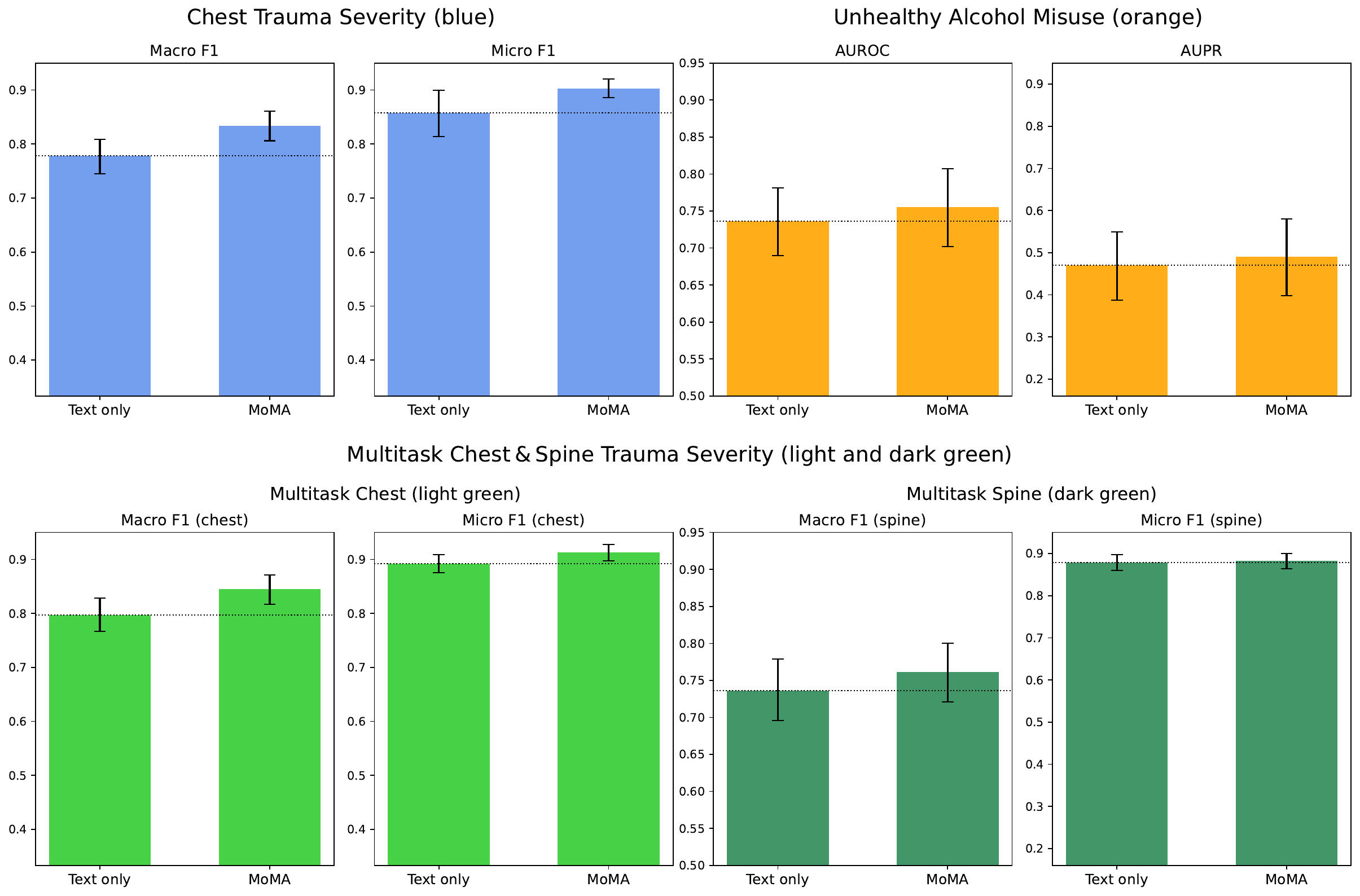}
    \caption{Ablation study evaluating the impact of removing non-text modalities from MoMA's input. MoMA consistently achieves superior performance compared to its text-only counterparts, demonstrating that performance improvements arise from the effective utilization of non-text modalities.}
    \label{fig:ablation}
\end{figure}

To validate the contribution of non-text modalities to the improved performance, we conducted ablation studies in which non-text inputs were removed while keeping the remaining components of MoMA unchanged. Specifically, we excluded chest radiographs in the chest trauma severity stratification and the multitask chest and spine trauma severity stratification tasks and lab measurements in the unhealthy alcohol use screening cohort. The results are presented in Figure \ref{fig:ablation}, with corresponding numerical values provided in \ref{tab:ablation} for reference. MoMA with multimodal input outperformed its text-only counterparts. These results highlight that MoMA's improved performance is not only attributed to the enhanced text understanding capabilities of LLMs but also to the architecture's ability to effectively integrate and leverage non-text modalities.

\subsection*{Case Study}

\begin{figure}[ht!]
    \centering
    \includegraphics[width=0.8\textwidth]{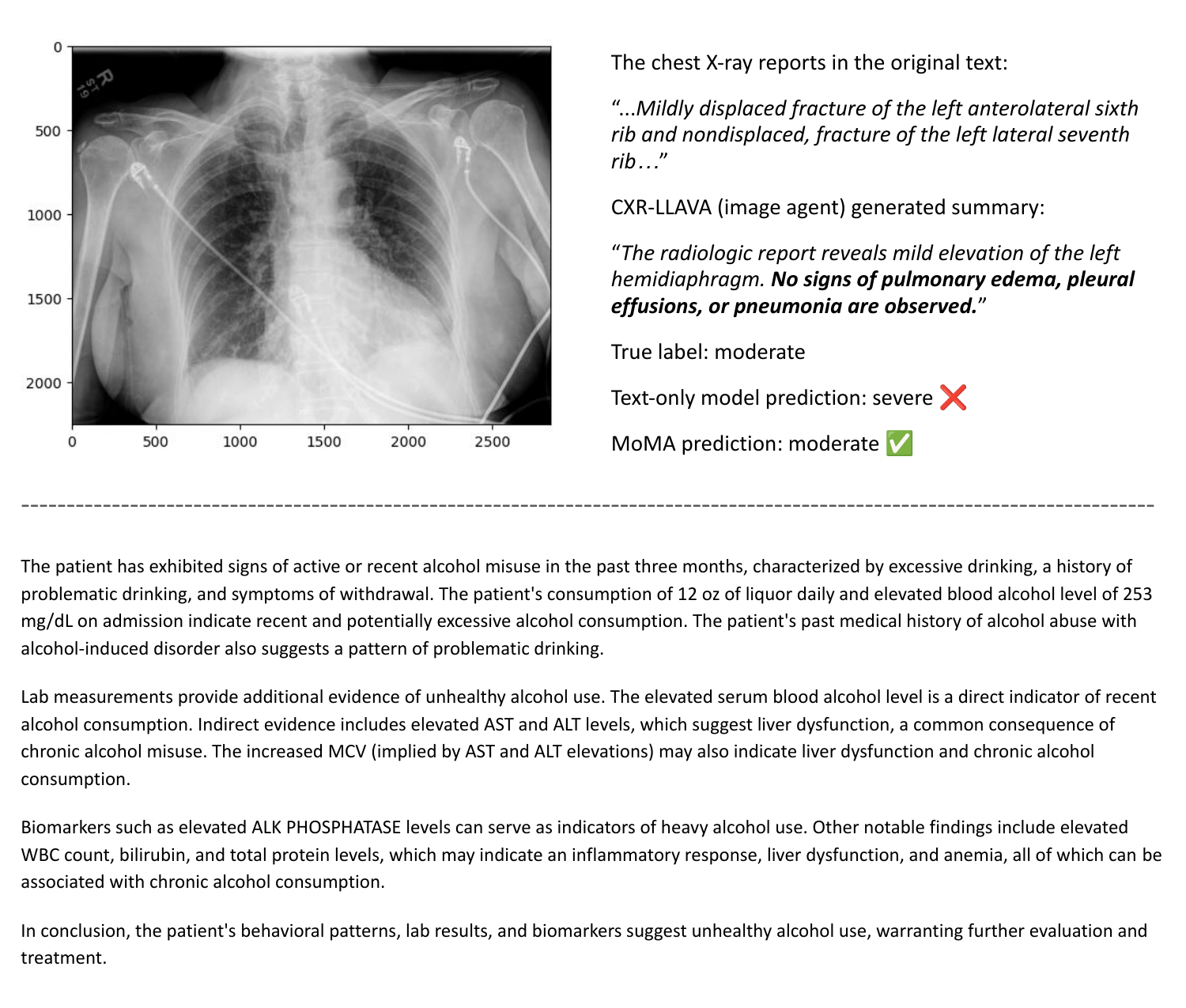}
    \caption{The upper section showcases the output of the chest X-ray specialist agent, which excludes the presence of severe conditions, enabling MoMA to accurately classify the case as a moderate injury. The lower section provides an example where clinical text and lab results are condensed into a concise summary by the specialist and aggregator agents. This summarized information allows the predictor agent to make predictions with improved interpretability compared to directly using all input text for classification.} 
    \label{fig:case}
\end{figure}

Figure \ref{fig:case} provides examples illustrating how the MoMA architecture effectively incorporated non-text information and synthesized all available data into an aggregated summary. In the upper example, the chest radiograph specialist agent analyzed radiographic images to identify or exclude severe conditions. In this case, the agent's output confirmed the absence of severe findings, enabling the architecture to accurately stratify the injury as moderate. However, the text-only approach incorrectly classified the case as severe, highlighting its limitation in utilizing critical radiographic information. In the lower example, the specialist agents and the aggregator agent collaboratively processed extensive clinical text (containing thousands of words) and lab measurements (comprising tens of measurements). These inputs were distilled into a concise, focused summary that preserved critical information while filtering out irrelevant details, which allowed the predictor agent to make an accurate classification while enhancing the transparency and interpretability of MoMA's decision-making process.

\section*{Discussion}

In this work, we introduced Mixture-of-Multimodal-Agents (MoMA), a flexible architecture designed to harness the power of pretrained LLMs for clinical prediction tasks involving multimodal medical data. MoMA adopts a modular, plug-and-play design that enables state-of-the-art LLMs to serve as specialist agents for processing specific data types, allowing them to be easily swapped or extended based on task requirements. The proposed architecture was validated on three clinical tasks involving different combinations of EHR modalities (radiographs + clinical text and lab measurements + clinical text) and different prediction types (multiclass, multitask, and binary classification). Additionally, all the tasks utilized private datasets, ensuring that there was no risk of data leakage from the LLMs' pretraining phase, as publicly available datasets may have been incorporated into their pretraining phase. Across these tasks, MoMA achieved superior performance compared to baseline methods, showcasing its potential as a highly effective and flexible solution to handling a wide range of clinical prediction tasks.

Current studies on multi-agent architectures have largely emphasized their enhanced text understanding capabilities, particularly in generative tasks \cite{jiang2024multi, du2023improving, tang2023medagents, li2024more}. However, they have yet to demonstrate whether such architectures can effectively leverage their text understanding abilities to improve prediction performance in classification tasks involving multimodal EHR data, which is a critical need in clinical applications. Our work directly addresses this gap by introducing MoMA, which harnesses the text understanding power of LLMs to achieve superior performance in clinical classification tasks across various modalities. To further illustrate its utility, we provided detailed case study results showcasing how MoMA’s specialist agents extract and summarize key information from diverse input modalities. These agents contribute unique perspectives, which are then integrated by the aggregator agent to distill complex data into concise summaries. This sequential approach improves the transparency of the decision-making process. In clinical settings, where understanding the rationale behind predictions is as important as achieving high accuracy, MoMA offers an effective solution that combines performance with interpretability.

Unlike common multimodal integration methodologies that require extensive paired pretraining data, MoMA offers a significant advantage in its flexibility to utilize existing pretrained models for ``projecting" non-text modalities into the text space. This independence allows MoMA to generalize better across diverse clinical tasks without being constrained by the availability of specific paired datasets during pretraining. By leveraging pretrained multimodal LLMs, MoMA encodes heterogeneous data types efficiently, combining their strengths in a unified architecture that is adaptable to varying input formats, such as sequential tabular data or variable numbers of medical images. Transforming all modalities to text also leverages the strengths of LLMs as language models that excel with textual input.

In this work, MoMA demonstrated superior performance in clinical tasks involving different combinations of EHR modalities: chest radiographs with clinical text, and lab measurements with clinical text. Leveraging recent advances in multimodal LLMs, MoMA can be readily adapted to various clinical scenarios by incorporating diverse non-text modalities beyond the chest radiographs and lab measurements presented here, simply by selecting appropriate specialist agents. For example, specialist agents such as BrainGPT\cite{li2025towards} for 3D CT scans, ConcepPath\cite{zhao2024aligning} for histopathology images, and scGPT\cite{cui2024scgpt} for single-cell sequencing data can be easily integrated into the MoMA workflow when these modalities are involved. Moreover, MoMA’s modular design facilitates the seamless integration of newer and more advanced LLMs as specialist, aggregator, or predictor agents, underscoring its flexibility and broad applicability.

In the chest trauma severity stratification and multitask chest and spine trauma severity stratification tasks, adding a third modality (e.g., lab measurement) to MoMA is feasible but did not result in performance improvement over the clinical text + radiographs setting. We attribute this to the already high performance achieved by the two modalities alone on these challenging multiclass tasks, suggesting that much of the relevant clinical signal may already be captured by these two rich modalities in this specific context. That said, MoMA supports the integration of different modalities with minimal overhead. This design align with our belief that clinical applications are inherently use-case specific: in some tasks, lab measurements may be critical; in others, they may be redundant or less informative.

Because our evaluation used private datasets, external researchers cannot directly access the original data. To address this and enhance reproducibility, we also demonstrated MoMA using publicly available datasets (MIMIC-IV\cite{johnson2023mimic} and MIMIC-CXR-JPG\cite{johnson2019mimic}), including admission notes and chest radiograph images, to predict in-hospital mortality in patients admitted to the Trauma Surgical Intensive Care Unit (TSICU). Instructions for reproducing MoMA’s performance are available via the link provided in the ``Code Availability'' section.

Another key advantage of MoMA is that it doesn't require any training process for the non-text modality translation. By requiring fine-tuning only on the predictor agent, MoMA reduces the computational burden and data requirements typically associated with training large-scale multimodal models. This not only accelerates the training process but also makes MoMA more accessible for applications with limited resources, further demonstrating its versatility.

Hallucination \cite{huang2025survey} is a known limitation of LLMs. While directly addressing this issue is beyond the primary scope of our study and remains an open challenge in the field, we have taken steps to mitigate its potential impact within our classification pipeline. MoMA does not rely on the generated summaries for human interpretation. Instead, these summaries serve as intermediate representations used for downstream prediction. The final classifications are produced by feedforward layers applied to the state embeddings of the predictor agent, and the predictor agent is fine-tuned using groundtruth clinical labels. This training process encourages alignment between model predictions and true clinical outcomes, thereby reducing the negative impact from hallucinations that may arise in the generated summaries.

While MoMA shows promise for low resource with high accuracy predictions, several limitations need further exploration. First, the interactions between LLM agents in MoMA remain relatively simple, and enhancing communication and coordination of agents \cite{fourney2024magentic} could further improve the model’s capabilities. Second, although MoMA offers flexibility in selecting pretrained LLMs in a plug-and-play manner, users should be aware that limitations such as introducing hallucination \cite{huang2025survey} and omitting necessary signals \cite{caffagni2024revolution} are inherent to LLMs and may still impact classification performance, though MoMA is fine-tuned using groundtruth clinical labels. Finally, while this study focuses on clinical classification tasks, MoMA can be potentially extended to broader applications, such as medical visual question answering (Medical VQA) \cite{ben2019vqa}; however, further validation is needed to support such extensions.

In summary, MoMA represents an advancement in leveraging the strength of LLMs with multimodal medical data for clinical prediction tasks. It demonstrates superior performance compared to state-of-the-art methods, meanwhile offering interpretability, computational efficiency, and flexibility to diverse input formats, making it a highly promising tool for improving clinical decision making.




\section*{Methods}
\label{sec:methods}

In this section, we present our proposed MoMA architecture. We begin with outlining the foundational design insights behind MoMA. Then we describe the roles of each agent and their collaborative processes in generating the final output in detail. In addition, we introduce the Cross-attention and Mixture-of-Experts based fusion methods which serve as comparison baselines. We adhere to the Transparent Reporting of a Multivariable Prediction Model for Individual Prognosis or Diagnosis - Large Language Models (TRIPOD-LLM) guidelines \cite{gallifant2025tripod} and completed the accompanying checklist, meeting all reporting requirements, as detailed in \ref{appendix:tripod}. 

\subsection*{Mixture-of-Multimodal-Agents Architecture}
\label{subsec:moma}

The design of the MoMA architecture draws upon two frameworks: Mixture-of-Experts (MoE) and Mixture-of-Agents (MoA). The Mixture-of-Experts (MoE) framework \cite{shazeer2017outrageously} is a well-established approach for enhancing model performance by selectively activating specialized ``experts". The basic mechanism behind MoE involves a gating network that learns to route each input to the most relevant experts, thus enabling the model to adaptively leverage specialist knowledge without overloading each expert with irrelevant information. The MoE approach has demonstrated excellent performance across a range of tasks within natural language processing \cite{muennighoff2022mteb} and computer vision domains.

Building on this foundation, the Mixture-of-Agents (MoA) architecture \cite{wang2024mixture} extends the MoE concept by replacing traditional experts and the gating network with LLM agents, thereby leveraging the distinct expertise of different pretrained LLMs. It also harnesses the collaborativeness of LLMs, wherein an LLM agent typically achieves better performance when it incorporates the output of another LLM agent as additional input, compared to relying solely on the original input text. The MoA framework achieves state-of-the-art performance in text understanding and summarization tasks.

In this study, we extend the MoA framework to accommodate multiple modalities in EHR data, proposing the MoMA architecture. Unlike the original MoA which employs multiple LLM agents to analyze a single text input, MoMA assigns a dedicated specialist agent to convert each non-text modality into a text summary. These summaries from various modalities are then processed through a stack of LLMs to leverage their collaborative capabilities. This architecture can support a growing number of modalities through the integration of multimodal LLMs in a plug-and-play manner. Building upon this foundational insight, we now describe the specific details of the MoMA architecture.

Figure \ref{fig:architecture}  illustrates the architecture of MoMA. As one of the core design principles of MoMA is to align various modalities to the text space, the original clinical text remains unprocessed until it reaches the aggregator agent. Specialist agents convert each non-text modality (e.g., images, structured lab results) into concise textual summaries. For example, multimodal LLMs such as LLaVA-Med \cite{li2024llava} and CXR-LLAVA \cite{lee2023cxr} can generate text summaries from medical images, while structured EHR data can be summarized using general-purpose LLMs \cite{shi2024ehragent} like Llama-3 \cite{dubey2024llama}. The resulting text summaries from non-text modalities are concatenated with the original clinical notes, and this concatenated text is then provided as input to the aggregator agent. Formally, for sample $i$, the input to the aggregator agent is

\[
m_i = t_i \oplus \left( \bigoplus_{j=1}^{M} \mathcal{S}_j(k_{i,j}) \right)
\]


where $t_i$ is the original clinical text of sample $i$, $\oplus$ denotes text concatenation, $M$ denotes number of non-text modalities, $\bigoplus_{j=1}^{M}$ represents the operation of text concatenation across the summaries generated from the $M$ non-text modalities, $\mathcal{S}_j$ is the specialist agent handling the $j$th non-text modality, $k_{i, j}$ is the input data for the $j$th non-text modality of sample $i$.

The aggregator agent receives $m_i$ as input and is prompted to generate a comprehensive and concise summary that integrates information from all available modalities. This summary is then passed to the predictor agent, which uses it to make predictions. Formally, these steps are represented as follows:
\[
\hat{y}_i = \mathcal{P}(\mathcal{A} (m_i))
\]

where $\mathcal{A}$ is the aggregator agent, $\mathcal{P}$ is the predictor agent, and $\hat{y}_i$ is the output prediction of sample $i$. In our study, we utilize Llama-3 as the specialist agent for both free text and tabular data, while CXR-LLAVA serves as the specialist agent for chest radiographs. Additionally, Llama-3 is used as both the aggregator and predictor agent. To generate the final prediction, we extract the hidden state corresponding to the last token output by the aggregator agent (fine-tuned Llama-3) and pass it through a feedforward layer to produce the final logit. 

\subsection*{Prompt Engineering for MoMA}

We provide general guidelines for creating prompts tailored to the MoMA framework. Users can input the elements listed below into an LLM to automatically generate these prompts. Specifically, prompts should guide the text specialist agent to extract task-specific clinical information, instruct the non-text specialist agent to identify data that is both relevant to and complementary to clinical text, and direct the aggregator agent to synthesize and summarize outputs from all specialist agents. Example prompts for chest trauma severity stratification and unhealthy alcohol use screening are provided in \ref{sec:prompts}.

Outlined below are the prompting guidelines for the text specialist agent.

\begin{enumerate}
    \item \textbf{Identify Relevant Points}
    \begin{itemize}
        \item Read the entire note carefully. Highlight any direct mentions related to your primary focus.
        \item \emph{Example: ``As an experienced trauma physician...summarizing chest trauma injuries.''}
    \end{itemize}

    \item \textbf{Apply Specified Criteria}
    \begin{itemize}
        \item Compare the details you’ve noted to any stated thresholds (e.g., a certain number of drinks per day, specific radiology report findings).
        \item \emph{Example: ``Excessive Consumption: four or more drinks in a single day if female, five or more if male...''}
    \end{itemize}

    \item \textbf{Incorporate Additional Patient Attributes}
    \begin{itemize}
        \item Look for coexisting conditions or factors that might explain or overlap with your primary findings (e.g., other diagnoses, lifestyle issues).
        \item \emph{Example: ``Scan the note for conditions...like hepatitis, renal dysfunction...''}
    \end{itemize}

    \item \textbf{Maintain Clarity and Flow}
    \begin{itemize}
        \item Present the findings in one coherent summary. Start with definitive evidence, then mention borderline or uncertain items, and end with confounders.
        \item \emph{Example: ``Summarize your findings by clearly separating evidence of [X] from evidence of other conditions...''}
    \end{itemize}

    \item \textbf{Professional Language and Confidentiality}
    \begin{itemize}
        \item Use objective, clinical terminology
        \item Omit Protected Health Information (PHI) and avoid speculation.
        \item \emph{Example: ``Do not include any Protected Health Information (PHI)... Do not guess or infer based on information that does not directly prove it.''}
    \end{itemize}

    \item \textbf{Final Structured Review}
    \begin{itemize}
        \item Conclude with a succinct overview of confirmed evidence, partial/absent findings, and relevant confounders.
        \item \emph{Example:``If no direct evidence meeting these criteria is found, explicitly state that no direct evidence is found.''}
    \end{itemize}

\end{enumerate}

The following outlines the prompting guidelines for the non-text specialist agent.

\begin{enumerate}
    \item \textbf{Identify Relevant Indicators}
    \begin{itemize}
        \item Scan for direct measures relevant to the outcome (e.g., blood alcohol levels).
        \item Look for key values linked to the condition or focus of your review.
        \item \emph{Example: ``Identify any initial measurements commonly linked to alcohol consumption...''}
    \end{itemize}

    \item \textbf{Evaluate Indirect Evidence}
    \begin{itemize}
        \item Check secondary or indirect measures relevant to the outcome (e.g., abnormal enzyme levels).
        \item Note whether these indirectly suggest the issue at hand.
        \item \emph{Example: ``Consider labs with indirect evidence... elevated liver enzymes...''}
    \end{itemize}

    \item \textbf{Summarize Concisely}
    \begin{itemize}
        \item Create a short, clear overview highlighting the most pertinent findings.
        \item Maintain professionalism and exclude any unnecessary speculation.
    \end{itemize}
\end{enumerate}

The guidelines for prompting the aggregator specialist agent are provided below.

\begin{enumerate}
    \item \textbf{Gather Key Points from Agent-Generated Summaries}
    \begin{itemize}
        \item Collect all relevant findings from each summary, such as clinical observations, lab findings, or imaging results.
        \item \emph{Example: ``Review the agent-generated summaries to identify any details related to alcohol use, including behavior patterns...''}
    \end{itemize}

    \item \textbf{Handle Contradicted Information}
    \begin{itemize}
        \item If the reports are contradictory, do not let automatically generated information overwrite established confirmed evidence.
        \item \emph{Example: ``Ensure that the LLM-generated radiology reports do not override clinical notes in cases where they contradict each other...''}
    \end{itemize}

    \item \textbf{Exclude Confounding Information}
    \begin{itemize}
        \item Exclude any details if they have an alternative cause not relevant to the goal.
        \item \emph{Example: ``For any mentioned lab abnormalities, review the clinical summaries to determine if they may have causes unrelated to alcohol use.''}
    \end{itemize}

    \item \textbf{Create a Unified Summary}
    \begin{itemize}
        \item Generate a comprehensive summary by integrating agent-generated reports from multimodal data for specific prediction tasks.
        \item \emph{Example: ``Write a comprehensive summary of the patient’s alcohol use, integrating relevant details from both the clinical summaries and lab results.''}
    \end{itemize}

\end{enumerate}

\subsection*{LLaVA-Med}
We adopt the LLaVA-Med v1.5 (Mistral-7B) checkpoint trained in April 2024 as a medical vision-language large model benchmark \cite{li2024llava}. This model extends the general-purpose LLaVA framework to the medical domain by adapting it for radiograph interpretation and multimodal instruction following.

LLaVA-Med integrates a CLIP image encoder \cite{radford2021learning} with a Mistral-7B language model \cite{jiang2023mistral7b} via a learned projection layer that aligns visual embeddings to the language model’s input space. This model extends the general-purpose LLaVA framework \cite{liu2023visual} to the medical domain by adapting it for radiograph interpretation and multimodal instruction following.

Following the classification setup used in the original LLaVA-Med paper, we treat the prediction task as a generative process. Specifically, we prompt the model with a clinical query and ask it to choose from a predefined list of candidate class labels (e.g., negative, moderate, severe). The final prediction is determined by string matching of the model’s response.

To address the known performance degradation with long or noisy textual input for medical vision-language large models \cite{xia2024mmed, xia2024rule}, we followed the strategy of Thawkar et al.\cite{thawkar2023xraygpt}, using a separate large language model to first summarize the full clinical notes into concise descriptions of trauma-related information. The resulting summary is then paired with the chest X-ray image and provided to LLaVA-Med for fine-tuning.

We fine-tuned LLaVA-Med on the development set to enhance predictive performance. During training, we supervised the model using paired inputs (chest radiographs + summarized clinical notes) and corresponding class labels as instruction–response pairs. The prompts used for this benchmark have been presented in \ref{sec:prompts}.

\subsection*{Cross-attention Based Fusion}
\label{subsec:cross-attention}
Cross-attention provides a robust framework for fusing encoded representations from two modalities by allowing one modality to selectively attend to relevant features in the other. Let $U_a$ and $U_b$ denote the encoded representations of two modalities, where $U_a$ provides the primary signals for classification, and $U_b$ serves as the complementary modality. The cross-attention mechanism computes a refined representation of $U_a$ by attending to $U_b$ as follows.

It first conducts linear transformations that project $U_a$ and $U_b$ into query ($Q$), key ($K$), and value ($V$) spaces:
\begin{align*}
Q &= U_a W_Q, \\
K &= U_b W_K, \\
V &= U_b W_V
\end{align*}
where $W_Q, W_K, W_V$ are learnable weight matrices.

Then scaled dot-product attention computes the alignment scores between $Q$ and $K$:
\[
A = \text{softmax}\left(\frac{Q K^\top}{\sqrt{d}}\right),
\]
where $A$ is the attention weight matrix, normalized row-wise to ensure row-wise weights sum to 1.

The values $V$ are aggregated based on the attention weights to produce the fused representation $F_a$:
\[
F_a = A V,
\]
where $F_a$ is the updated representation of $U_a$, enriched with relevant information from $U_b$. The refined representation $F_a$ is combined with the original $U_a$ via residual connection to form the final fused representation for downstream tasks.
\[
F = F_a + U_a.
\]

\subsection*{Sparse Mixture-of-Experts (MoE) Based Fusion}
\label{subsec:moe}
The Sparse Mixture-of-Experts (MoE) model is an efficient approach for fusing encoded representations from $M$ different EHR modalities, denoted by $\{U_1, U_2, \dots, U_M\}$, including both text and non-text modalities. The process is outlined as follows:

The MoE framework includes a set of $K$ expert networks $\{E_1, E_2, \dots, E_K\}$. Each expert specializes in processing subsets of the input space, enabling the model to handle diverse patterns across modalities. Each expert processes one or more encoded representations $U_m$ based on the routing mechanism. A gating network determines how to route the input representations $U_m$ to the experts. The gating network generates scores $G_{m,k}$ for each combination of modality $m$ and expert $k$:

\[
G_{m,k} = \text{softmax}(\phi(U_m)),
\]
where $\phi$ is a neural network applied to $U_m$.

Then the outputs from the active experts are aggregated to form the final fused representation $F$. 

\[
F = \sum_{k \in \text{active}} G_{m,k} \cdot E_k(U_m).
\]

The fused representation $F$ is passed to a final prediction layer to perform the downstream classification task.

\subsection*{Published SOTA}
\label{subsec:sota}
Following the design proposed by Gao et al.\cite{gao2024automated}, we fine-tuned ClinicalBERT using clinical text inputs, specifically ED notes and radiology reports. Due to ClinicalBERT’s input length limit, notes that are more likely to provide comprehensive information were prioritized: ED notes were sorted from longest to shortest, and radiology reports from earliest to latest. The first 300 tokens were allocated to ED notes, with the remaining tokens assigned to radiology reports. If one note type used fewer tokens than its allocation, the unused tokens were reallocated to the other. Notes exceeding the overall token limit were truncated. This ClinicalBERT baseline was finetuned on the development set of this work and validated on the test set independently. We used this baseline for the chest trauma severity and the multitask chest and spine trauma severity stratification tasks.

In the work by Afshar et al.\cite{afshar2022development}, clinical concepts, such as diseases, symptoms, anatomical sites, medications, and procedures, were extracted from all available notes within each patient encounter using the NLP engine, clinical Text And Knowledge Extraction System (cTAKES)\cite{savova2010mayo}. These extracted concepts were then embedded and processed by a one-dimensional convolutional neural network (1D-CNN) for multi-task predictions of alcohol, opioid, and non-opioid misuse. We utilized the trained model from \cite{afshar2022development} and reported the performance of its unhealthy alcohol use prediction head on our test set without any additional fine-tuning. In addition, we retrained the model on the same cohort used by MoMA for the unhealthy alcohol use screening task.

\subsection*{Computational Resources and Runtime}
All experiments were conducted using two A100 GPUs with 80 GB of memory. For the chest trauma severity stratification task and the multitask chest and spine trauma severity stratification task, generating summaries with specialist agents takes approximately 72 hours, while the remaining processes are completed in under 4 hours. For the unhealthy alcohol use screening task, specialist summary generation took around 48 hours, with the remaining processes completed in under 3 hours.

\subsection*{Ethics Statement}
This study was approved by the University of Wisconsin-Madison Minimal Risk Research Institutional Review Board (IRB) under the following protocols. For the chest trauma severity stratification and multitask chest and spine trauma severity stratification tasks, ethical approval was granted under IRB protocol \#2019-1258 with a waiver of informed consent. For the unhealthy alcohol use screening task, ethical approval was granted under IRB protocol \#2021-0509. Informed consent was obtained from all participants.

\section*{Data Availability}
Due to legal and regulatory constraints, the data utilized in this study are not publicly accessible. Our data was obtained from the UW Health system after receiving approval from the IRB. Our data use agreements do not permit sharing clinical data. Researchers with an interest in accessing the data can reach out to the corresponding authors or Madeline Oguss at mkoguss@medicine.wisc.edu. A demonstration of MoMA, using the publicly available MIMIC-IV and MIMIC-CXR datasets, is provided to ensure the reproducibility of our methods.

The code for this project, including the chest trauma severity stratification and unhealthy alcohol use screening tasks as well as the demonstration using the MIMIC-IV and MIMIC-CXR datasets, is available at \url{https://git.doit.wisc.edu/smph-public/dom/uw-icu-data-science-lab-public/moma}

\section*{Acknowledgments}
This was supported in part by the National Institute of Health/National Institute on Drug Abuse R01DA051464 (MA, MO), National Library of Medicine R01LM012973 (MA) and the National Science Foundation DMS-2054346 (JG, GC).

\section*{Author Contributions}
G. C. and M. A. conceived the study; J. G. developed the methods with input from G. C. and M.A.; J. G. conducted the simulations and application with help from J. C.; A.R., R. B., A. S., A. M. assisted result interpretation; G. C., M.A. and M. C. supervised the study; G. C. and M.A. provided funding, J. G., G. C. and M. A. wrote the draft. All authors reviewed and provided revision input on the manuscript.

\section*{Conflicts of Interest}
The authors have no competing interests to declare.

\printbibliography[segment=2,title={References}]
\end{refsegment}


\clearpage 

\begin{table}[!htbp]
\centering
\caption{Cohort characteristics and label distributions. The columns under \textit{``Trauma severity stratification"} present statistics of both the chest trauma severity stratification and the multitask chest and spine trauma severity stratification task, as these two tasks were conducted on the same cohort.}
\begin{tabular}{*5l}
\toprule
{}   & \multicolumn{2}{c}{Trauma severity stratification} & \multicolumn{2}{c}{Unhealthy alcohol use screening}  \\
{}   &  Development &  Test &  Development &  Test \\
\midrule
Encounters, n & 3,451 & 870 & 1,576 & 482 \\
Age, median (IQR) &  58 (37,74) & 62 (42,77) & 63 (50,73) & 64 (51,73)\\
Total hours of stay, median (IQR) & 104 (59,189) & 104 (52,190) & 100 (65,175) & 144 (84,253)\\
Female, n (\%) & 1,303 (37.8) & 346 (39.8) & 744 (47.2) & 207 (42.9)\\
White, n (\%) & 3,180 (92.1) & 786 (90.3) & 1,418 (90.0) & 432 (89.6)\\
\midrule
\multicolumn{5}{l}{Labels for chest trauma severity stratification} \\
\quad\quad Negative & 2,184 (63.3) & 538 (61.8) & - & -\\
\quad\quad Minor/moderate & 398 (11.5) & 97 (11.1) & - & -\\
\quad\quad Serious or greater & 869 (25.1) & 235 (27.0) & - & -\\
\midrule
\multicolumn{5}{l}{Labels for spine trauma severity stratification} \\
\quad\quad Negative & 2,394 (69.4) & 626 (72.0) & - & -\\
\quad\quad Minor/moderate & 773 (22.4) & 187 (21.5) & - & -\\
\quad\quad Serious or greater & 284 (8.2) & 57 (6.6) & - & -\\
\midrule
\multicolumn{5}{l}{Labels for unhealthy alcohol use screening} \\
\quad\quad Negative & - & - & 235 (14.9) & 78 (16.2)\\
\quad\quad Positive &- & - & 1,341 (85.1) & 404 (83.8)\\
\bottomrule
\end{tabular}
\label{tab:cohort_char}
\end{table}

\clearpage

\clearpage 

\newpage

\begin{refsection}
\setcounter{page}{1}
\appendix
\section*{Supplementary Material}

\setcounter{table}{0}
\renewcommand{\thetable}{Supplementary Table \arabic{table}} 
\setcounter{section}{0}
\renewcommand{\thesubsection}{Supplementary Note \arabic{subsection}}
\renewcommand{\tablename}{}

\subsection{Additional Evaluation Results} \label{appendix:main_results} 
\vspace{-3mm} \noindent \ref{tab: chest_alcohol_results} and \ref{tab: multitask_results} present primary metrics (macro/micro-F1 for multiclass classification tasks, AUROC and AUPR for binary classification tasks). \ref{tab: alternative_evaluation} shows alternative metrics (macro-AUROC for multiclass classification tasks, F1 for binary classification tasks). \ref{tab: trauma_subgroup}, \ref{tab: multitask_chest_subgroup}, \ref{tab: multitask_spine_subgroup} and \ref{tab: alcohol_subgroup} present subgroup analyses of MoMA and baselines. \ref{tab:ablation} and \ref{tab:multitask_ablation} show ablation studies evaluating when removing non-text components from MoMA. \ref{tab: 1d_cnn-sample_size} present the performance comparison of the published baselines (1D-CNN) for unhealthy alcohol use screening trained with different sample sizes. \ref{tab:third_modality} compares the results when adding a third modality (lab measurements) to MoMA in the chest trauma and multitask trauma severity stratification tasks.

\begin{table}[!htbp]
\centering
\caption{Discrimination performance on the test set for chest trauma severity stratification and unhealthy alcohol use screening}
\begin{tabular}{l*5c}
\toprule
{}   & \multicolumn{2}{c}{\textit{Chest trauma severity stratification}} & \multicolumn{2}{c}{\textit{Unhealthy alcohol use screening}}  \\
{}  & macro-F1 & micro-F1 & AUROC & AUPR \\
\midrule
Published baseline & \makecell{0.636\\(0.609,0.663)} & \makecell{0.692\\(0.666,0.717)} & \makecell{0.714\\(0.655,0.771)} & \makecell{0.431\\(0.339,0.517)} \\
Cross-attention fusion & \makecell{0.632\\(0.601,0.663)} & \makecell{0.770\\(0.747,0.793)} & \makecell{0.647\\(0.581,0.709)} & \makecell{0.339\\(0.255,0.424)} \\
MoE fusion & \makecell{0.630\\(0.605,0.654)} & \makecell{0.756\\(0.730,0.785)} & \makecell{0.670\\(0.610,0.729)} & \makecell{0.374\\(0.284,0.467)} \\
LLaVA-Med & \makecell{0.802\\(0.775,0.829)} & \makecell{0.883\\(0.868,0.899)} & - & - \\
MoMA & \makecell{0.834\\(0.806,0.861)} & \makecell{0.903\\(0.886,0.920)} & \makecell{0.755\\(0.702,0.807)} & \makecell{0.491\\(0.398,0.580)} \\
\bottomrule
\end{tabular}
\label{tab: chest_alcohol_results}
\end{table}

\begin{table}[!htbp]
\centering
\caption{Discrimination performance on the test set for multitask chest and spine trauma severity stratification}
\begin{tabular}{l*5c}
\toprule
{}   & \multicolumn{2}{c}{\textit{Multitask Chest}} & \multicolumn{2}{c}{\textit{Multitask Spine}}  \\
{}  & macro-F1 & micro-F1 & macro-F1 & micro-F1 \\
\midrule
Published baseline & \makecell{0.630\\(0.599,0.662)} & \makecell{0.794\\(0.771,0.816)} & \makecell{0.494\\(0.476,0.513)} & \makecell{0.781\\(0.757,0.803)} \\
Cross-attention fusion & \makecell{0.621\\(0.585,0.648)} & \makecell{0.784\\(0.759,0.806)} & \makecell{0.495\\(0.476,0.512)} & \makecell{0.783\\(0.761,0.805)} \\
MoE fusion & \makecell{0.609\\(0.581,0.647)} & \makecell{0.777\\(0.750,0.803)} & \makecell{0.486\\(0.464,0.511)} & \makecell{0.775\\(0.749,0.801)} \\
LLaVA-Med & \makecell{0.806\\(0.773,0.832)} & \makecell{0.889\\(0.870,0.906)} & \makecell{0.694\\(0.653,0.735)} & \makecell{0.856\\(0.837,0.876)} \\
MoMA & \makecell{0.845\\(0.817,0.871)} & \makecell{0.913\\(0.897,0.928)} & \makecell{0.761\\(0.721,0.800)} & \makecell{0.883\\(0.864,0.900)} \\
\bottomrule
\end{tabular}
\label{tab: multitask_results}
\end{table}

\begin{table}[!htbp]
\centering
\caption{Alternative metrics}
\begin{tabular}{l*5c}
\toprule
{}   & \multicolumn{1}{c}{\textit{\makecell{Chest trauma\\severity stratification} }} & \multicolumn{1}{c}{\textit{\makecell{Unhealthy\\alcohol use screening}}} & \multicolumn{2}{c}{\textit{\makecell{Multitask trauma\\severity stratification}}} \\
{}  & macro-AUROC & F1 & \makecell{macro-AUROC\\(chest)} & \makecell{macro-AUROC\\(spine)} \\
\midrule
Published baseline & \makecell{0.826\\(0.806,0.844)} & \makecell{0.400\\(0.332,0.474)} & \makecell{0.865\\(0.847,0.883)} & \makecell{0.844\\(0.822,0.864)} \\
Cross-attention fusion & \makecell{0.815\\(0.797,0.831)} & \makecell{0.403\\(0.344,0.478)} & \makecell{0.851\\(0.830,0.869)} & \makecell{0.836\\(0.807,0.851)} \\
MoE fusion & \makecell{0.814\\(0.800,0.829)} & \makecell{0.423\\(0.364,0.503)} & \makecell{0.849\\(0.827,0.866)} & \makecell{0.832\\(0.804,0.846)} \\
LLaVA-Med & \makecell{0.953\\(0.937,0.968)} & \makecell{-} & \makecell{0.956\\(0.940,0.969)} & \makecell{0.910\\(0.894,0.935)} \\
MoMA & \makecell{0.960\\(0.945,0.973)} & \makecell{0.478\\(0.385,0.558)} & \makecell{0.958\\(0.947,0.968)} & \makecell{0.935\\(0.920,0.948)} \\
\bottomrule
\end{tabular}
\label{tab: alternative_evaluation}
\end{table}

\vspace{-1mm}
\begin{table}[!htbp]
\centering
\caption{Subgroup Analysis for Chest Trauma Severity Stratification}
\adjustbox{scale=1.15,center}{
\resizebox{\textwidth}{!}{
\begin{tabular}{|l|cccc|cccc|}
\hhline{|---------|}
{}   & \multicolumn{4}{c|}{\textit{macro-F1}} & \multicolumn{4}{c|}{\textit{micro-F1}} \\
\hhline{|~~~~~~~~~|}
{}  & Male & Female & White & Non-white & Male & Female & White & Non-white \\
\hhline{|---------|}
Published baseline & \makecell{0.599\\(0.564,0.634)} & \makecell{0.696\\(0.564,0.634)} & \makecell{0.638\\(0.610,0.667)} & \makecell{0.612\\(0.524,0.702)} & \makecell{0.646\\(0.612,0.681)} & \makecell{0.760\\(0.723,0.798)} & \makecell{0.695\\(0.668,0.722)} & \makecell{0.656\\(0.571,0.738)} \\
\hhline{|---------|}
Cross-attention fusion & \makecell{0.622\\(0.583,0.661)} & \makecell{0.642\\(0.590,0.694)} & \makecell{0.636\\(0.604,0.669)} & \makecell{0.586\\(0.496,0.680)} & \makecell{0.755\\(0.725,0.786)} & \makecell{0.792\\(0.754,0.829)} & \makecell{0.777\\(0.752,0.801)} & \makecell{0.703\\(0.619,0.786)} \\
\hhline{|---------|}
MoE fusion & \makecell{0.625\\(0.575,0.658)} & \makecell{0.638\\(0.601,0.672)} & \makecell{0.631\\(0.610,0.653)} & \makecell{0.570\\(0.509,0.640)} & \makecell{0.731\\(0.709,0.754)} & \makecell{0.779\\(0.748,0.810)} & \makecell{0.766\\(0.741,0.790)} & \makecell{0.702\\(0.647,0.755)}\\
\hhline{|---------|}
LLaVA-Med & \makecell{0.800\\(0.762,0.837)} & \makecell{0.819\\(0.769,0.863)} & \makecell{0.812\\(0.780,0.843)} & \makecell{0.761\\(0.666,0.854)} & \makecell{0.878\\(0.853,0.901)} & \makecell{0.908\\(0.882,0.934)} & \makecell{0.895\\(0.877,0.913)} & \makecell{0.833\\(0.762,0.893)}\\
\hhline{|---------|}
MoMA & \makecell{0.828\\(0.790,0.864)} & \makecell{0.845\\(0.799,0.888)} & \makecell{0.841\\(0.810,0.870)} & \makecell{0.819\\(0.756,0.868)} & \makecell{0.893\\(0.870,0.916)} & \makecell{0.919\\(0.893,0.942)} & \makecell{0.910\\(0.892,0.926)} & \makecell{0.886\\(0.841,0.934)}\\
\hhline{|---------|}
\end{tabular}
}
}
\label{tab: trauma_subgroup}
\end{table}

\vspace{-1mm}
\begin{table}[!htbp]
\centering
\caption{Subgroup Analysis for \textbf{Multitask} Chest Trauma Severity Stratification}
\adjustbox{scale=1.15,center}{
\resizebox{\textwidth}{!}{
\begin{tabular}{|l|cccc|cccc|}
\hhline{|---------|}
{}   & \multicolumn{4}{c|}{\textit{macro-F1}} & \multicolumn{4}{c|}{\textit{micro-F1}} \\
\hhline{|~~~~~~~~~|}
{}  & Male & Female & White & Non-white & Male & Female & White & Non-white \\
\hhline{|---------|}
Published baseline & 
\makecell{0.605\\(0.568,0.644)} & \makecell{0.667\\(0.615,0.720)} & \makecell{0.629\\(0.595,0.663)} & \makecell{0.621\\(0.528,0.717)} & \makecell{0.757\\(0.727,0.786)} & \makecell{0.849\\(0.818,0.883)} & \makecell{0.794\\(0.770,0.817)} & \makecell{0.797\\(0.726,0.869)} \\
\hhline{|---------|}
Cross-attention fusion &
\makecell{0.571\\(0.528,0.600)}& \makecell{0.611\\(0.565,0.659)}& \makecell{0.585\\(0.545,0.621)}& \makecell{0.579\\(0.502,0.655)}& \makecell{0.743\\(0.714,0.771)}& \makecell{0.839\\(0.799,0.878)}& \makecell{0.783\\(0.742,0.820)}& \makecell{0.781\\(0.709,0.867)}\\
\hhline{|---------|}
MoEfusion &
\makecell{0.575\\(0.531,0.617)}& \makecell{0.610\\(0.568,0.673)}& \makecell{0.584\\(0.546,0.622)}& \makecell{0.578\\(0.505,0.660)}& \makecell{0.760\\(0.731,0.789)}& \makecell{0.855\\(0.821,0.886)}& \makecell{0.800\\(0.775,0.826)}& \makecell{0.785\\(0.701,0.862)}\\
\hhline{|---------|}
LLaVA-Med &
\makecell{0.784\\(0.745,0.823)}& \makecell{0.833\\(0.785,0.879)}& \makecell{0.807\\(0.775,0.838)}& \makecell{0.763\\(0.668,0.853)}& \makecell{0.870\\(0.844,0.895)}& \makecell{0.916\\(0.890,0.939)}& \makecell{0.893\\(0.874,0.911)}& \makecell{0.844\\(0.774,0.905)}\\
\hhline{|---------|}
MoMA &
\makecell{0.851\\(0.816,0.884)}& \makecell{0.834\\(0.788,0.879)}& \makecell{0.840\\(0.810,0.869)}& \makecell{0.861\\(0.785,0.931)}& \makecell{0.908\\(0.887,0.929)}& \makecell{0.919\\(0.896,0.942)}& \makecell{0.915\\(0.898,0.930)}& \makecell{0.892\\(0.833,0.940)}\\
\hhline{|---------|}
\end{tabular}
}
}
\label{tab: multitask_chest_subgroup}
\end{table}

\vspace{-1mm}
\begin{table}[!htbp]
\centering
\caption{Subgroup Analysis for \textbf{Multitask} Spine Trauma Severity Stratification}
\adjustbox{scale=1.15,center}{
\resizebox{\textwidth}{!}{
\begin{tabular}{|l|cccc|cccc|}
\hhline{|---------|}
{}   & \multicolumn{4}{c|}{\textit{macro-F1}} & \multicolumn{4}{c|}{\textit{micro-F1}} \\
\hhline{|~~~~~~~~~|}
{}  & Male & Female & White & Non-white & Male & Female & White & Non-white \\
\hhline{|---------|}
Published baseline &
\makecell{0.475\\(0.451,0.498)}& \makecell{0.528\\(0.498,0.557)}& \makecell{0.498\\(0.479,0.517)}& \makecell{0.452\\(0.380,0.520)}& \makecell{0.758\\(0.727,0.788)}& \makecell{0.815\\(0.780,0.850)}& \makecell{0.784\\(0.758,0.808)}& \makecell{0.750\\(0.679,0.821)}\\
\hhline{|---------|}
Cross-attention fusion &
\makecell{0.471\\(0.445,0.499)}& \makecell{0.472\\(0.441,0.504)}& \makecell{0.485\\(0.456,0.517)}& \makecell{0.430\\(0.341,0.509)}& \makecell{0.766\\(0.731,0.799)}& \makecell{0.816\\(0.782,0.847)}& \makecell{0.785\\(0.760,0.805)}& \makecell{0.755\\(0.683,0.826)}\\
\hhline{|---------|}
MoEfusion &
\makecell{0.480\\(0.460,0.501)}& \makecell{0.519\\(0.490,0.552)}& \makecell{0.492\\(0.473,0.511)}& \makecell{0.443\\(0.351,0.524)}& \makecell{0.755\\(0.721,0.781)}& \makecell{0.820\\(0.779,0.858)}& \makecell{0.781\\(0.750,0.809)}& \makecell{0.750\\(0.674,0.823)}\\
\hhline{|---------|}
LLaVA-Med &
\makecell{0.705\\(0.656,0.753)}& \makecell{0.652\\(0.574,0.729)}& \makecell{0.704\\(0.660,0.747)}& \makecell{0.571\\(0.435,0.733)}& \makecell{0.839\\(0.813,0.865)}& \makecell{0.881\\(0.853,0.910)}& \makecell{0.859\\(0.838,0.879)}& \makecell{0.832\\(0.762,0.893)}\\
\hhline{|---------|}
MoMA &
\makecell{0.792\\(0.747,0.835)}& \makecell{0.749\\(0.672,0.828)}& \makecell{0.762\\(0.720,0.802)}& \makecell{0.738\\(0.568,0.878)}& \makecell{0.878\\(0.855,0.901)}& \makecell{0.890\\(0.860,0.916)}& \makecell{0.882\\(0.863,0.899)}& \makecell{0.874\\(0.820,0.931)}\\
\hhline{|---------|}
\end{tabular}
}
}
\label{tab: multitask_spine_subgroup}
\end{table}

\vspace{-1mm}
\begin{table}[!htbp]
\centering
\caption{Subgroup Analysis for Unhealthy Alcohol Use Screening}
\begin{tabular}{l*5c}
\toprule
{}   & \multicolumn{2}{c}{\textit{AUROC}} & \multicolumn{2}{c}{\textit{AUPR}}  \\
{}  & Male & Female & Male & Female \\
\midrule
Published baseline & \makecell{0.733\\(0.661,0.803)} & \makecell{0.682\\(0.583,0.774)} & \makecell{0.491\\(0.376,0.597)} & \makecell{0.346\\(0.210,0.479)} \\
Cross-attention fusion & \makecell{0.681\\(0.602,0.757)} & \makecell{0.585\\(0.476,0.691)} & \makecell{0.378\\(0.276,0.490)} & \makecell{0.329\\(0.191,0.471)} \\
MoE fusion & \makecell{0.667\\(0.590,0.742)} & \makecell{0.674\\(0.571,0.770)} & \makecell{0.398\\(0.284,0.516)} & \makecell{0.364\\(0.220,0.515)} \\
MoMA & \makecell{0.782\\(0.719,0.839)} & \makecell{0.745\\(0.659,0.838)} & \makecell{0.543\\(0.429,0.648)} & \makecell{0.468\\(0.316,0.616)} \\
\bottomrule
\end{tabular}
\label{tab: alcohol_subgroup}
\end{table}

\begin{table}[!htbp]
\centering
\caption{Ablation study}
\begin{tabular}{l*5c}
\toprule
{}   & \multicolumn{2}{c}{\textit{Chest trauma severity stratification}} & \multicolumn{2}{c}{\textit{Unhealthy alcohol use screening}}  \\
{}  & macro-F1 & micro-F1 & AUROC & AUPR \\
\midrule
Non-text modality removed & \makecell{0.778\\(0.745,0.809)} & \makecell{0.858\\(0.814,0.900)} & \makecell{0.736\\(0.690,0.781)} & \makecell{0.470\\(0.387,0.549)} \\
MoMA & \makecell{0.834\\(0.806,0.861)} & \makecell{0.903\\(0.886,0.920)} & \makecell{0.755\\(0.702,0.807)} & \makecell{0.491\\(0.398,0.580)} \\
\bottomrule
\end{tabular}
\label{tab:ablation}
\end{table}

\begin{table}[!htbp]
\centering
\caption{Ablation study}
\begin{tabular}{l*5c}
\toprule
{}   & \multicolumn{2}{c}{\textit{Multitask Chest}} & \multicolumn{2}{c}{\textit{Multitask Spine}}  \\
{}  & macro-F1 & micro-F1 & macro-F1 & micro-F1 \\
\midrule
Non-text modality removed & \makecell{0.797\\(0.767,0.828)} & \makecell{0.892\\(0.875,0.909)}  & \makecell{0.736\\(0.696,0.779)}  & \makecell{0.878\\(0.860,0.897)}  \\
MoMA & \makecell{0.845\\(0.817,0.871)} & \makecell{0.913\\(0.897,0.928)}  & \makecell{0.761\\(0.721,0.800)}  & \makecell{0.883\\(0.864,0.900)} \\
\bottomrule
\end{tabular}
\label{tab:multitask_ablation}
\end{table}

\begin{table}[!htbp]
\centering
\caption{Comparison of the published baselines (1D-CNN) for unhealthy alcohol use screening using different training data}
\begin{tabular}{l*3c}
\toprule
{}  & AUROC & AUPR \\
\midrule
Published model (sample size: 54,915) & \makecell{0.714\\(0.655,0.771)} & \makecell{0.431\\(0.339,0.517)} \\
Retrained model (sample size: 1,576) & \makecell{0.641\\(0.566,0.719)} & \makecell{0.325\\(0.231,0.419)} \\
\bottomrule
\end{tabular}
\label{tab: 1d_cnn-sample_size}
\end{table}

\begin{table}[!htbp]
\centering
\caption{Comparison of MoMA with two (text+radiographs) and three modalities (text+radiographs+labs) combined for chest trauma and multitask trauma severity stratification}
\adjustbox{scale=1.10,center}{
\resizebox{\textwidth}{!}{
\begin{tabular}{l*7c}
\toprule
{}   & \multicolumn{2}{c}{\textit{Chest Trauma}} & \multicolumn{2}{c}{\textit{Multitask Chest}} & \multicolumn{2}{c}{\textit{Multitask Spine}}  \\
{}  & macro-F1 & micro-F1 & macro-F1 & micro-F1 & macro-F1 & micro-F1 \\
\midrule
\makecell{Two Modalities\\(text+radiographs)} & \makecell{0.834\\(0.806,0.861)} & \makecell{0.903\\(0.886,0.920)} & \makecell{0.845\\(0.817,0.871)} & \makecell{0.913\\(0.897,0.928)}  & \makecell{0.761\\(0.721,0.800)}  & \makecell{0.883\\(0.864,0.900)} \\
\makecell{Three Modalities\\(text+radiographs+labs)}  & \makecell{0.836\\(0.805,0.863)} & \makecell{0.899\\(0.885,0.922)} & \makecell{0.841\\(0.815,0.869)} & \makecell{0.909\\(0.893,0.925)}  & \makecell{0.765\\(0.724,0.805)}  & \makecell{0.891\\(0.869,0.904)} \\
\bottomrule
\end{tabular}
}}
\label{tab:third_modality}
\end{table}

\clearpage
\subsection{Optimization Hyperparameters}
\label{sec:technical_details}

\ref{tab:config_chest}, \ref{tab:config_alcohol}, and \ref{tab:config_multitask} are the technical details of developing MoMA for the three evaluation tasks. Full scripts and configs are available in our GitLab repository at \url{https://git.doit.wisc.edu/smph-public/dom/uw-icu-data-science-lab-public/moma}

\begin{table}[h]
\centering
\caption{Default hyperparameters for chest trauma severity stratification}
\begin{tabular}{lc}
\toprule
\textbf{Parameter} & \textbf{Value}  \\
\midrule
Specialist agent & CXR-LLAVA-v2 \\
Aggergator agent & Llama-3 8B \\
Predictor agent & Llama-3 8B \\
Optimizer & AdamW  \\
Learning rate & $1\!\times\!10^{-4}$ \\
Load in 8bit & True  \\
LoRA alpha & 16 \\
LoRA dropout & 0.05 \\
LoRA rank & 128 \\
Batch size & 2 \\
Max steps & 4500 \\
Warm-up steps & 2 \\
Gradient accumulation steps & 1 \\
Loss function & categorical cross-entropy \\
\bottomrule
\end{tabular}
\label{tab:config_chest}
\end{table}

\begin{table}[h]
\centering
\caption{Default hyperparameters for multitask chest and spine trauma severity stratification}
\begin{tabular}{lc}
\toprule
\textbf{Parameter} & \textbf{Value}  \\
\midrule
Specialist agent & CXR-LLAVA-v2 \\
Aggergator agent & Llama-3 8B \\
Predictor agent & Llama-3 8B \\
Optimizer & AdamW  \\
Learning rate & $1\!\times\!10^{-4}$ \\
Load in 8bit & True  \\
LoRA alpha & 16 \\
LoRA dropout & 0.05 \\
LoRA rank & 128 \\
Batch size & 2 \\
Max steps & 3500 \\
Warm-up steps & 2 \\
Gradient accumulation steps & 1 \\
Loss function & Total categorical cross-entropy of the two sub-tasks\\
\bottomrule
\end{tabular}
\label{tab:config_alcohol}
\end{table}

\begin{table}[h]
\centering
\caption{Default hyperparameters for unhealthy alcohol use screening}
\begin{tabular}{lc}
\toprule
\textbf{Parameter} & \textbf{Value}  \\
\midrule
Specialist agent & Llama-3 8B \\
Aggergator agent & Llama-3 8B \\
Predictor agent & Llama-3 8B \\
Optimizer & AdamW  \\
Learning rate & $1\!\times\!10^{-4}$ \\
Load in 8bit & True  \\
LoRA alpha & 32 \\
LoRA dropout & 0.05 \\
LoRA rank & 32 \\
Batch size & 2 \\
Max steps & 4500 \\
Weight decay & 0.01 \\
Warm-up steps & 2 \\
Gradient accumulation steps & 1 \\
Loss function & binary cross-entropy with logits\\
\bottomrule
\end{tabular}
\label{tab:config_multitask}
\end{table}

\newpage

\subsection{Prompts}
\label{sec:prompts}
\subsubsection{Prompt for the chest X-ray specialist agent in the chest trauma severity stratification task}
We did not provide prompts to CXR-LLAVA \cite{lee2023cxr}, the chest X-ray specialist agent, as it performs effectively without additional prompting.

\subsubsection{Prompt for the aggregator agent in the chest trauma severity stratification task}

\begin{verbatim}
As an experienced trauma physician, your task is to review the clinical notes, radiology
reports, and LLM-generated radiology reports. Write a summary focusing on identifying and
summarizing chest trauma injuries, and determine the chest Abbreviated Injury Scale (AIS).

Follow these steps to complete the task:

	1. Extract and summarize information related to the severity of chest trauma from the provided 
    clinical notes and radiology reports. Do not include injuries in body regions outside the chest.
    
	2. If the reports of X-RAY CHEST AP/PA/Single VIEW are not available, summarize the LLM-generated 
    radiology reports as complementary information. Ensure that the LLM-generated reports do 
    not overwrite clinical notes if they contradict each other.
    
	3. Based on your summary, determine the chest AIS score (ranging from 0 to 6).
    Ensure the assessment is exclusively based on trauma-related conditions/symptoms.
    
	4. Provide your conclusion as a single-digit number ranging from 0 to 6.
\end{verbatim}

\subsubsection{Prompt for the aggregator agent in the multitask chest and spine trauma severity stratification}

\begin{verbatim}
As an experienced trauma physician, your task is to review the clinical notes, radiology
reports, and LLM-generated radiology reports. Write a summary focusing on identifying and
summarizing chest and spine trauma injuries, and determine the chest Abbreviated Injury Scale (AIS).

Follow these steps to complete the task:

	1. Extract and summarize information related to the severity of only chest 
    and spine trauma from the provided 
    clinical notes and radiology reports. Do not include injuries in body regions
    outside chest and spine.
    
	2. If the reports of X-RAY CHEST AP/PA/Single VIEW are not available, summarize
    the LLM-generated radiology reports as complementary information. Ensure that 
    the LLM-generated reports do not overwrite clinical notes if they contradict each other.
    
	3. Based only on trauma-related conditions or symptoms, assign an Abbreviated 
    Injury Scale (AIS) score (0–6) for each region.
    Remember that only conditions/symptoms caused by trauma injuries should be 
    used to determine the AIS scores.
    
	4. Based on the AIS scores of chest and spine injuries, translate to the Severity
    Category for chest and spine:
    AIS = 0 → Negative
    AIS = 1 or AIS = 2 → Moderate
    AIS > 2 → Serious
\end{verbatim}

\subsubsection{Prompt for the lab measurement specialist agent in the unhealthy alcohol use screening task}

\begin{verbatim}
As an expert in screening for unhealthy alcohol use, carefully review the provided lab 
measurements and generate a concise summary highlighting potential indicators of unhealthy 
alcohol use based on your analysis. Let's think through this step by step:
1. Identify any initial measurements commonly linked to alcohol consumption or misuse like serum 
blood alcohol levels.
2. Consider labs with indirect evidence for unhealthy alcohol consumption (e.g., elevated liver
enzymes, mean corpuscular volume).
3. Incorporate labs that have previously been shown to serve as biomarkers of unhealthy alcohol 
use.

Make sure your response is short and concise. Avoid being verbose.

Here are a few examples:
Direct Indicator: The serum blood alcohol level of 12 mg/dL is above the legal 
limit and indicates recent alcohol consumption.
Indirect Evidence: Elevated AST and ALT levels, along with an increased MCV, suggest liver 
dysfunction and macrocytosis, both of which are commonly associated with chronic alcohol misuse.
Biomarker: The GGT level is significantly elevated, which can serve as a biomarker for 
heavy alcohol use.

\end{verbatim}

\subsubsection{Prompt for the aggregator agent in the unhealthy alcohol use screening task}

\begin{verbatim}
Role:
You are an alcohol screener working within a healthcare system, responsible for 
determining whether a patient has exhibited signs of unhealthy alcohol use over the 
past three months. Your evaluation will be based on clinical summaries generated by 
LLM agents, which include clinical notes and lab measurements.

Objective:
Develop a focused summary of the patient’s alcohol use. Ensure that no personally 
identifiable information (PHI) is included.

Task Instructions:

1. Assess Evidence of Alcohol Misuse:
	- Review the clinical summaries to identify any details related to alcohol use, 
    including behavioral patterns, attempts to manage drinking, or external concerns.
	- Focus on direct evidence and ensure the summary highlights key findings related to 
    alcohol use while avoiding unnecessary or redundant information.

2. Summarize Lab Measurements:
	- Pay close attention to direct evidence from lab results, such as blood alcohol 
    concentration (BAC) levels, as they provide clear indications of alcohol use.
	- Include other lab abnormalities only if they are explicitly connected to alcohol use.

3. Evaluate Causes of Lab Abnormalities:
	- For any mentioned lab abnormalities, review the clinical summaries to determine if 
    they may have causes unrelated to alcohol use.
	- Exclude such lab results from the summary and explicitly state when a lab abnormality
    have an alternative cause.

4. Compose a Unified Summary:
	- Write a comprehensive summary of the patient’s alcohol use, integrating relevant 
    details from both the clinical summaries and lab results.
	- Ensure the summary prioritizes key findings, focusing primarily on direct evidence
    such as BAC levels and behavioral indications of alcohol use.

\end{verbatim}

\subsubsection{Prompt for summarizing clinical notes for LLaVA-Med in the multitask chest and spine trauma severity stratification}

\begin{verbatim}
You are a clinical summarization assistant.
Your job is to read the given ED notes and radiology reports, then extract only
the details related to chest trauma and spine trauma, separately.

1. Produce two labeled sections in your response:
 - Chest Trauma Summary: 
 - Spine Trauma Summary: 

2. Keep each summary short and self-contained. Do not mention or quote which 
section(s) of the note the information came from.

3. If no chest trauma is mentioned, exactly reply:
> No chest trauma mentioned in the clinical note.
   If no spine trauma is mentioned, exactly reply:
> No spine trauma mentioned in the clinical note.
   If neither chest nor spine trauma is mentioned, exactly reply:
> No chest or spine trauma mentioned in the clinical note.

4. Do not include any additional commentary or information beyond the two summaries
or one of the exact “No … mentioned” statements.
\end{verbatim}

\subsubsection{Prompt for summarizing clinical notes for LLaVA-Med in the chest trauma severity stratification}

\begin{verbatim}
You are a clinical summarization assistant.
Your job is to read the given ED notes and radiology reports, then extract only the
details related to chest trauma.

1. Keep the summary short and self-contained. Do not mention or quote which 
section(s) of the note the information came from.

3. If no chest trauma is mentioned, exactly reply:
> No chest trauma mentioned in the clinical note.

4. Do not include any additional commentary or information beyond the summary or the
exact “No … mentioned” statements.
\end{verbatim}

\subsubsection{Prompt for generating severity predictions for LLaVA-Med in the chest trauma severity stratification}

\begin{verbatim}
You are a radiology assistant specialized in chest trauma.
Given a chest X-ray and a brief clinical note summary,
classify the trauma severity on a scale from:
0 = no trauma
1 = minor or moderate trauma
2 = serious or greater than serious trauma
Reply with exactly one integer (like '1,2').

\end{verbatim}

\subsubsection{Prompt for generating severity predictions for LLaVA-Med in the multitask chest and spine trauma severity stratification}

\begin{verbatim}
You are a radiology assistant specialized in chest and spine trauma.
Given a chest X-ray and a brief clinical note summary,
classify the trauma severity on a scale from:
0 = no trauma
1 = minor or moderate trauma
2 = serious or greater than serious trauma
Reply with exactly two integers separated by comma (like '1,2'), one for chest 
and one for spine, and no other text.

\end{verbatim}

\subsection{TRIPOD-LLM checklist}
\label{appendix:tripod}

The TRIPOD-LLM \cite{gallifant2025tripod} checklist is attached below. Based on the instructions, we selected ``\textbf{LLM methods}" as the research design and ``\textbf{Classification}" as the research task. All the relevant items are reported below in \ref{tab:TRIPOD-LLM}

\begin{longtable}{lp{9cm}p{1.8cm}}
        \toprule
        
        \textbf{\emph{Section / Topic}} & \textbf{\emph{Checklist Item}} & \textbf{\emph{Reported on Page}} \\ 
        \midrule
        
        \textbf{Abstract} & {} & {} \\
        \midrule
        Title & Identify the study as developing, fine-tuning, and/or evaluating the performance of an LLM, specifying the task, the target population, and the outcome to be predicted. & 1 \\ 
        \midrule
        Objective & Specify the study objectives, including whether the study describes LLMs development, tuning, and/or evaluation & 1 \\ 
        \midrule
        \multirow{6}{*}{Methods} & Describe the key elements of the study setting. & 1\\ 
        & Detail all data used in the study, specify data splits and any selective use of data. & 1\\ 
        & Specify the name and version of LLM used. & 7\\ 
        & Briefly summarize the LLM-building steps, including any fine-tuning, reward modeling, reinforcement learning with human feedback (RLHF), etc. & 1 \\ 
        & Describe the specific tasks performed by the LLMs (e.g., medical QA, summarization, extraction), highlighting key inputs and outputs used in the final LLM. & 1 \\ 
        & Specify the evaluation datasets/populations used, including the endpoint evaluated, and detail whether this information was held out during training/tuning where relevant, and what measure(s) were used to evaluate LLM performance. & 1 \\ 
        \midrule
        {Results} & Give an overall report and interpretation of the main results. & 2,3 \\ 
        \midrule
        {Discussion} & Explicitly state any broader implications or concerns that have arisen in light of these results. & 2,3 \\ 
        \midrule
        
        \textbf{Introduction} & {} & {} \\
        {Background} & Explain the healthcare context / use case (e.g., administrative, diagnostic, therapeutic, clinical workflow) and rationale for developing or evaluating the LLM, including references to existing approaches and models. & 1,2,3 \\ 
        \midrule
        {Objectives} & Specify the study objectives, including whether the study describes the initial development, fine-tuning, or validation of an LLM (or multiple stages). & 1,2,3 \\ 
        \midrule
        
        \textbf{Methods} & {} & {} \\
        \midrule
        \multirow{5}{*}{Data} & Describe the sources of data separately for the training, tuning, and/or evaluation datasets and the rationale for using these data (e.g., web corpora, clinical research/trial data, EHR data). & 3 \\ 
        & Describe the relevant data points and provide a quantitative and qualitative description of their distribution and other relevant descriptors of the dataset (e.g., source, languages, countries of origin) & 3,15 \\ 
        & Specifically state the date of the oldest and newest item of text used in the development process (training, fine-tuning, reward modeling) and in the evaluation datasets.  & 3,15 \\ 
        & Describe any data pre-processing and quality checking, including whether this was similar across text corpora, institutions, and relevant sociodemographic groups. & 3 \\ 
        &Describe how missing and imbalanced data were handled and provide reasons for omitting any data.  & 3  \\ 
        \midrule
        \multirow{5}{*}{Analytical Methods} & Report the LLM name, version, and last date of training or use during inference. & 7 \\ 
        &Specify the type of LLM architecture, and LLM building steps, including any hyperparameter tuning (e.g., temperature, length limits, penalties), prompt engineering, and any inference settings (e.g., seed, temperature, max token length) as relevant.  &6,7,8,9,10 \\
        &Report details of LLM development process from text input to outcome generation, such as training, fine-tuning procedures, and alignment strategy (e.g., reinforcement learning, direct preference optimization, etc.) and alignment goals (e.g., helpfulness, honesty, harmlessness, etc.).  &6,7,8,9,10  \\ 
        &Specify the initial and post-processed output of the LLM (e.g., probabilities, classification, unstructured text).  &6,7,8,9,10  \\
        &Provide details and rationale for any classification and how the probabilities were determined and thresholds identified.  &3  \\ 
        \midrule
        \multirow{2}{*}{LLM Output} & If outcome assessment requires subjective interpretation, describe the qualifications of the assessors, any instructions provided, relevant information on demographics of the assessors, and inter-assessor agreement. & 3 \\ 
         & Specify how performance was compared to other LLMs, humans, and other benchmarks or standards.  & 4,5,16,17,18 \\ 
         \midrule
        \multirow{3}{*}{Annotation} & If annotation was done, report how text was labeled, including providing specific annotation guidelines with examples. & 3 \\ 
        & If annotation was done, report how many annotators labeled the dataset(s), including the proportion of data in each dataset that were annotated by more than 1 annotator.  & 3 \\ 
        & If annotation was done, provide information on the background and experience of the annotators, and the inter-annotator agreement.  & 3  \\
        \midrule
        \multirow{2}{*}{Prompting} & If research involved prompting LLMs, provide details on the processes used during prompt design, curation, and selection. & 7,8,9 \\ 
        & If research involved prompting LLMs, report what data were used to develop the prompts.  & 7,8,9 \\ 
        \midrule
        Instruction Tuning / Alignment & If instruction tuning/alignment strategies were used, what were the instructions and interface used for evaluation, and what were the characteristics of the populations doing evaluation? & N/A. Instruction tuning/alignment strategies are not used. \\ 
        \midrule
        Compute & Report compute, or proxies thereof (e.g., time on what and how many machines, cost on what and how many machines, inference time, floating-point operations per second (FLOPs)), required to carry out methods. & 10 \\ 
        \midrule
        Ethics Approval & Name the institutional research board or ethics committee that approved the study and describe the participant-informed consent or the ethics committee waiver of informed consent.& 10 \\ 
        \midrule
        \multirow{4}{*}{Open Science} &Give the source of funding and the role of the funders for the present study.  &14 \\
        &Declare any conflicts of interest and financial disclosures for all authors.  &14  \\ 
        &Provide details of the availability of the study data.  &14\\
        &Provide details of the availability of the code to reproduce the study results. &14  \\ 
        \midrule
        
        \textbf{Results} & {} & {} \\
        \midrule
        Performance & Report LLM performance according to pre-specified metrics (see item 7a) and/or human evaluation (see item 7d). & 3,4,15,16,17,18 \\ 
        \midrule
        LLM Updating & If applicable, report the results from any LLM updating, including the updated LLM and subsequent performance. & 3,4,15,16,17,18 \\ 
        \midrule
        
        \textbf{Discussion} & {} & {} \\
        \midrule
        Interpretation & Give an overall interpretation of the main results, including issues of fairness in the context of the objectives and previous studies. & 5,6\\ 
        \midrule
        Limitations & Discuss any limitations of the study and their effects on any biases, statistical uncertainty, and generalizability. & 6 \\ 
        \midrule
        Usability of the LLM in context & Discuss any next steps for future research, with a specific view to applicability and generalizability of the LLM. & 6 \\ 
        \bottomrule
    \caption{TRIPOD-LLM checklist}
    \label{tab:TRIPOD-LLM}
\end{longtable}

\printbibliography[heading=subbibliography,title={References for Supplementary Material}]

@article{cai2019survey,
  title={A survey on multimodal data-driven smart healthcare systems: approaches and applications},
  author={Cai, Qiong and Wang, Hao and Li, Zhenmin and Liu, Xiao},
  journal={IEEE Access},
  volume={7},
  pages={133583--133599},
  year={2019},
  publisher={IEEE}
}

@article{gao2024automated,
  title={Automated stratification of trauma injury severity across multiple body regions using multi-modal, multi-class machine learning models},
  author={Gao, Jifan and Chen, Guanhua and O’Rourke, Ann P and Caskey, John and Carey, Kyle A and Oguss, Madeline and Stey, Anne and Dligach, Dmitriy and Miller, Timothy and Mayampurath, Anoop and others},
  journal={Journal of the American Medical Informatics Association},
  volume={31},
  number={6},
  pages={1291--1302},
  year={2024},
  publisher={Oxford University Press}
}

@article{kline2022multimodal,
  title={Multimodal machine learning in precision health: A scoping review},
  author={Kline, Adrienne and Wang, Hanyin and Li, Yikuan and Dennis, Saya and Hutch, Meghan and Xu, Zhenxing and Wang, Fei and Cheng, Feixiong and Luo, Yuan},
  journal={npj Digital Medicine},
  volume={5},
  number={1},
  pages={171},
  year={2022},
  publisher={Nature Publishing Group UK London}
}

@article{rohaut2024multimodal,
  title={Multimodal assessment improves neuroprognosis performance in clinically unresponsive critical-care patients with brain injury},
  author={Rohaut, Benjamin and Calligaris, C and Hermann, B and Perez, P and Faugeras, F and Raimondo, F and King, J- R and Engemann, D and Marois, C and Le Guennec, L and others},
  journal={Nature Medicine},
  pages={1--7},
  year={2024},
  publisher={Nature Publishing Group US New York}
}

@article{soenksen2022integrated,
  title={Integrated multimodal artificial intelligence framework for healthcare applications},
  author={Soenksen, Luis R and Ma, Yu and Zeng, Cynthia and Boussioux, Leonard and Villalobos Carballo, Kimberly and Na, Liangyuan and Wiberg, Holly M and Li, Michael L and Fuentes, Ignacio and Bertsimas, Dimitris},
  journal={NPJ digital medicine},
  volume={5},
  number={1},
  pages={149},
  year={2022},
  publisher={Nature Publishing Group UK London}
}

@inproceedings{winston2024multimodal,
  title={Multimodal Clinical Prediction with Unified Prompts and Pretrained Large-Language Models},
  author={Winston, Caleb and Winston, Chloe and Winston, Cailin and Winston, Claris and Winston, Cleah},
  booktitle={2024 IEEE 12th International Conference on Healthcare Informatics (ICHI)},
  pages={679--683},
  year={2024},
  organization={IEEE}
}

@article{acosta2022multimodal,
  title={Multimodal biomedical AI},
  author={Acosta, Juli{\'a}n N and Falcone, Guido J and Rajpurkar, Pranav and Topol, Eric J},
  journal={Nature Medicine},
  volume={28},
  number={9},
  pages={1773--1784},
  year={2022},
  publisher={Nature Publishing Group US New York}
}

@article{stahlschmidt2022multimodal,
  title={Multimodal deep learning for biomedical data fusion: a review},
  author={Stahlschmidt, S{\"o}ren Richard and Ulfenborg, Benjamin and Synnergren, Jane},
  journal={Briefings in Bioinformatics},
  volume={23},
  number={2},
  pages={bbab569},
  year={2022},
  publisher={Oxford University Press}
}

@article{huang2020fusion,
  title={Fusion of medical imaging and electronic health records using deep learning: a systematic review and implementation guidelines},
  author={Huang, Shih-Cheng and Pareek, Anuj and Seyyedi, Saeed and Banerjee, Imon and Lungren, Matthew P},
  journal={NPJ digital medicine},
  volume={3},
  number={1},
  pages={136},
  year={2020},
  publisher={Nature Publishing Group UK London}
}

@article{guarrasi2024systematic,
  title={A Systematic Review of Intermediate Fusion in Multimodal Deep Learning for Biomedical Applications},
  author={Guarrasi, Valerio and Aksu, Fatih and Caruso, Camillo Maria and Di Feola, Francesco and Rofena, Aurora and Ruffini, Filippo and Soda, Paolo},
  journal={arXiv preprint arXiv:2408.02686},
  year={2024}
}

@inproceedings{hayat2022medfuse,
  title={MedFuse: Multi-modal fusion with clinical time-series data and chest X-ray images},
  author={Hayat, Nasir and Geras, Krzysztof J and Shamout, Farah E},
  booktitle={Machine Learning for Healthcare Conference},
  pages={479--503},
  year={2022},
  organization={PMLR}
}

@inproceedings{radford2021learning,
  title={Learning transferable visual models from natural language supervision},
  author={Radford, Alec and Kim, Jong Wook and Hallacy, Chris and Ramesh, Aditya and Goh, Gabriel and Agarwal, Sandhini and Sastry, Girish and Askell, Amanda and Mishkin, Pamela and Clark, Jack and others},
  booktitle={International conference on machine learning},
  pages={8748--8763},
  year={2021},
  organization={PMLR}
}

@inproceedings{li2022blip,
  title={Blip: Bootstrapping language-image pre-training for unified vision-language understanding and generation},
  author={Li, Junnan and Li, Dongxu and Xiong, Caiming and Hoi, Steven},
  booktitle={International conference on machine learning},
  pages={12888--12900},
  year={2022},
  organization={PMLR}
}

@inproceedings{bannur2023learning,
  title={Learning to exploit temporal structure for biomedical vision-language processing},
  author={Bannur, Shruthi and Hyland, Stephanie and Liu, Qianchu and Perez-Garcia, Fernando and Ilse, Maximilian and Castro, Daniel C and Boecking, Benedikt and Sharma, Harshita and Bouzid, Kenza and Thieme, Anja and others},
  booktitle={Proceedings of the IEEE/CVF Conference on Computer Vision and Pattern Recognition},
  pages={15016--15027},
  year={2023}
}

@article{cohen2023joint,
  title={Joint variational autoencoders for multimodal imputation and embedding},
  author={Cohen Kalafut, Noah and Huang, Xiang and Wang, Daifeng},
  journal={Nature Machine Intelligence},
  volume={5},
  number={6},
  pages={631--642},
  year={2023},
  publisher={Nature Publishing Group UK London}
}

@article{zhao2023survey,
  title={A survey of large language models},
  author={Zhao, Wayne Xin and Zhou, Kun and Li, Junyi and Tang, Tianyi and Wang, Xiaolei and Hou, Yupeng and Min, Yingqian and Zhang, Beichen and Zhang, Junjie and Dong, Zican and others},
  journal={arXiv preprint arXiv:2303.18223},
  year={2023}
}

@article{wei2022emergent,
  title={Emergent abilities of large language models},
  author={Wei, Jason and Tay, Yi and Bommasani, Rishi and Raffel, Colin and Zoph, Barret and Borgeaud, Sebastian and Yogatama, Dani and Bosma, Maarten and Zhou, Denny and Metzler, Donald and others},
  journal={arXiv preprint arXiv:2206.07682},
  year={2022}
}

@article{thirunavukarasu2023large,
  title={Large language models in medicine},
  author={Thirunavukarasu, Arun James and Ting, Darren Shu Jeng and Elangovan, Kabilan and Gutierrez, Laura and Tan, Ting Fang and Ting, Daniel Shu Wei},
  journal={Nature medicine},
  volume={29},
  number={8},
  pages={1930--1940},
  year={2023},
  publisher={Nature Publishing Group US New York}
}

@article{wang2024drg,
  title={DRG-LLaMA: tuning LLaMA model to predict diagnosis-related group for hospitalized patients},
  author={Wang, Hanyin and Gao, Chufan and Dantona, Christopher and Hull, Bryan and Sun, Jimeng},
  journal={npj Digital Medicine},
  volume={7},
  number={1},
  pages={16},
  year={2024},
  publisher={Nature Publishing Group UK London}
}

@article{liu2023medical,
  title={A medical multimodal large language model for future pandemics},
  author={Liu, Fenglin and Zhu, Tingting and Wu, Xian and Yang, Bang and You, Chenyu and Wang, Chenyang and Lu, Lei and Liu, Zhangdaihong and Zheng, Yefeng and Sun, Xu and others},
  journal={NPJ Digital Medicine},
  volume={6},
  number={1},
  pages={226},
  year={2023},
  publisher={Nature Publishing Group UK London}
}

@article{gu2024probabilistic,
  title={Probabilistic Medical Predictions of Large Language Models},
  author={Gu, Bowen and Desai, Rishi J and Lin, Kueiyu Joshua and Yang, Jie},
  journal={arXiv preprint arXiv:2408.11316},
  year={2024}
}

@article{li2024llava,
  title={Llava-med: Training a large language-and-vision assistant for biomedicine in one day},
  author={Li, Chunyuan and Wong, Cliff and Zhang, Sheng and Usuyama, Naoto and Liu, Haotian and Yang, Jianwei and Naumann, Tristan and Poon, Hoifung and Gao, Jianfeng},
  journal={Advances in Neural Information Processing Systems},
  volume={36},
  year={2024}
}

@article{lee2023cxr,
  title={Cxr-llava: Multimodal large language model for interpreting chest x-ray images},
  author={Lee, Seowoo and Youn, Jiwon and Kim, Mansu and Yoon, Soon Ho},
  journal={arXiv preprint arXiv:2310.18341},
  year={2023}
}

@article{dubey2024llama,
  title={The llama 3 herd of models},
  author={Dubey, Abhimanyu and Jauhri, Abhinav and Pandey, Abhinav and Kadian, Abhishek and Al-Dahle, Ahmad and Letman, Aiesha and Mathur, Akhil and Schelten, Alan and Yang, Amy and Fan, Angela and others},
  journal={arXiv preprint arXiv:2407.21783},
  year={2024}
}

@article{zhu2024prompting,
  title={Prompting large language models for zero-shot clinical prediction with structured longitudinal electronic health record data},
  author={Zhu, Yinghao and Wang, Zixiang and Gao, Junyi and Tong, Yuning and An, Jingkun and Liao, Weibin and Harrison, Ewen M and Ma, Liantao and Pan, Chengwei},
  journal={arXiv preprint arXiv:2402.01713},
  year={2024}
}

@article{gao2024raw,
  title={When Raw Data Prevails: Are Large Language Model Embeddings Effective in Numerical Data Representation for Medical Machine Learning Applications?},
  author={Gao, Yanjun and Myers, Skatje and Chen, Shan and Dligach, Dmitriy and Miller, Timothy A and Bitterman, Danielle and Churpek, Matthew and Afshar, Majid},
  journal={arXiv preprint arXiv:2408.11854},
  year={2024}
}

@article{wang2024mixture,
  title={Mixture-of-Agents Enhances Large Language Model Capabilities},
  author={Wang, Junlin and Wang, Jue and Athiwaratkun, Ben and Zhang, Ce and Zou, James},
  journal={arXiv preprint arXiv:2406.04692},
  year={2024}
}

@article{shazeer2017outrageously,
  title={Outrageously large neural networks: The sparsely-gated mixture-of-experts layer},
  author={Shazeer, Noam and Mirhoseini, Azalia and Maziarz, Krzysztof and Davis, Andy and Le, Quoc and Hinton, Geoffrey and Dean, Jeff},
  journal={arXiv preprint arXiv:1701.06538},
  year={2017}
}

@article{muennighoff2022mteb,
    doi = {10.48550/ARXIV.2210.07316},
    url = {https://arxiv.org/abs/2210.07316},
    author = {Muennighoff, Niklas and Tazi, Nouamane and Magne, Lo{\"\i}c and Reimers, Nils},
    title = {MTEB: Massive Text Embedding Benchmark},
    publisher = {arXiv},
    journal={arXiv preprint arXiv:2210.07316},  
    year = {2022}
}

@inproceedings{shi2024ehragent,
  title={Ehragent: Code empowers large language models for few-shot complex tabular reasoning on electronic health records},
  author={Shi, Wenqi and Xu, Ran and Zhuang, Yuchen and Yu, Yue and Zhang, Jieyu and Wu, Hang and Zhu, Yuanda and Ho, Joyce C and Yang, Carl and Wang, May Dongmei},
  booktitle={ICLR 2024 Workshop on Large Language Model (LLM) Agents},
  year={2024}
}

@article{jinyi2024global,
  title={Global, regional, and national mortality of tuberculosis attributable to alcohol and tobacco from 1990 to 2019: A modelling study based on the Global Burden of Disease study 2019},
  author={Jinyi, Wu and Zhang, Yue and Wang, Kai and Peng, Peng},
  journal={Journal of Global Health},
  volume={14},
  year={2024},
  publisher={International Society for Global Health}
}

@article{coulton2011alcohol,
  title={Alcohol misuse},
  author={Coulton, Simon},
  journal={BMJ Clinical Evidence},
  volume={2011},
  year={2011},
  publisher={BMJ Publishing Group}
}

@inproceedings{jiang2024multi,
  title={Multi-modal and multi-agent systems meet rationality: A survey},
  author={Jiang, Bowen and Xie, Yangxinyu and Wang, Xiaomeng and Su, Weijie J and Taylor, Camillo Jose and Mallick, Tanwi},
  booktitle={ICML 2024 Workshop on LLMs and Cognition},
  year={2024}
}

@article{du2023improving,
  title={Improving factuality and reasoning in language models through multiagent debate},
  author={Du, Yilun and Li, Shuang and Torralba, Antonio and Tenenbaum, Joshua B and Mordatch, Igor},
  journal={arXiv preprint arXiv:2305.14325},
  year={2023}
}

@article{li2024more,
  title={More agents is all you need},
  author={Li, Junyou and Zhang, Qin and Yu, Yangbin and Fu, Qiang and Ye, Deheng},
  journal={arXiv preprint arXiv:2402.05120},
  year={2024}
}

@article{fourney2024magentic,
  title={Magentic-one: A generalist multi-agent system for solving complex tasks},
  author={Fourney, Adam and Bansal, Gagan and Mozannar, Hussein and Tan, Cheng and Salinas, Eduardo and Niedtner, Friederike and Proebsting, Grace and Bassman, Griffin and Gerrits, Jack and Alber, Jacob and others},
  journal={arXiv preprint arXiv:2411.04468},
  year={2024}
}

@article{tang2023medagents,
  title={Medagents: Large language models as collaborators for zero-shot medical reasoning},
  author={Tang, Xiangru and Zou, Anni and Zhang, Zhuosheng and Li, Ziming and Zhao, Yilun and Zhang, Xingyao and Cohan, Arman and Gerstein, Mark},
  journal={arXiv preprint arXiv:2311.10537},
  year={2023}
}

@article{jian2024rethinking,
  title={Rethinking Cross-Attention for Infrared and Visible Image Fusion},
  author={Jian, Lihua and Xiong, Songlei and Yan, Han and Niu, Xiaoguang and Wu, Shaowu and Zhang, Di},
  journal={arXiv preprint arXiv:2401.11675},
  year={2024}
}

@article{zheng2024multimodal,
  title={Multimodal clinical trial outcome prediction with large language models},
  author={Zheng, Wenhao and Peng, Dongsheng and Xu, Hongxia and Li, Yun and Zhu, Hongtu and Fu, Tianfan and Yao, Huaxiu},
  journal={arXiv preprint arXiv:2402.06512},
  year={2024}
}

@article{herrera2022survival,
  title={From survival to survivorship—framing traumatic injury as a chronic condition},
  author={Herrera-Escobar, Juan P and Schneider, Jeffrey C},
  journal={The New England journal of medicine},
  volume={387},
  number={7},
  pages={581},
  year={2022},
  publisher={NIH Public Access}
}

@article{lefering2012epidemiology,
  title={Epidemiology of in-hospital trauma deaths},
  author={Lefering, R and Paffrath, T and Bouamra, O and Coats, TJ and Woodford, M and Jenks, T and Wafaisade, A and Nienaber, U and Lecky, F},
  journal={European journal of trauma and emergency surgery},
  volume={38},
  pages={3--9},
  year={2012},
  publisher={Springer}
}

@article{granstrom2018criteria,
  title={A criteria-directed protocol for in-hospital triage of trauma patients},
  author={Granstr{\"o}m, Anna and Str{\"o}mmer, Lovisa and Schandl, Anna and {\"O}stlund, Anders},
  journal={European Journal of Emergency Medicine},
  volume={25},
  number={1},
  pages={25--31},
  year={2018},
  publisher={LWW}
}

@article{afshar2022development,
  title={Development and multimodal validation of a substance misuse algorithm for referral to treatment using artificial intelligence (SMART-AI): a retrospective deep learning study},
  author={Afshar, Majid and Sharma, Brihat and Dligach, Dmitriy and Oguss, Madeline and Brown, Randall and Chhabra, Neeraj and Thompson, Hale M and Markossian, Talar and Joyce, Cara and Churpek, Matthew M and others},
  journal={The Lancet Digital Health},
  volume={4},
  number={6},
  pages={e426--e435},
  year={2022},
  publisher={Elsevier}
}

@article{han2024fusemoe,
  title={Fusemoe: Mixture-of-experts transformers for fleximodal fusion},
  author={Han, Xing and Nguyen, Huy and Harris, Carl and Ho, Nhat and Saria, Suchi},
  journal={arXiv preprint arXiv:2402.03226},
  year={2024}
}

@article{alsentzer2019publicly,
  title={Publicly available clinical BERT embeddings},
  author={Alsentzer, Emily and Murphy, John R and Boag, Willie and Weng, Wei-Hung and Jin, Di and Naumann, Tristan and McDermott, Matthew},
  journal={arXiv preprint arXiv:1904.03323},
  year={2019}
}

@article{mcneely2016performance,
  title={Performance of the tobacco, alcohol, prescription medication, and other substance use (TAPS) tool for substance use screening in primary care patients},
  author={McNeely, Jennifer and Wu, Li-Tzy and Subramaniam, Geetha and Sharma, Gaurav and Cathers, Lauretta A and Svikis, Dace and Sleiter, Luke and Russell, Linnea and Nordeck, Courtney and Sharma, Anjalee and others},
  journal={Annals of internal medicine},
  volume={165},
  number={10},
  pages={690--699},
  year={2016},
  publisher={American College of Physicians}
}

@article{li2023ethics,
  title={Ethics of large language models in medicine and medical research},
  author={Li, Hanzhou and Moon, John T and Purkayastha, Saptarshi and Celi, Leo Anthony and Trivedi, Hari and Gichoya, Judy W},
  journal={The Lancet Digital Health},
  volume={5},
  number={6},
  pages={e333--e335},
  year={2023},
  publisher={Elsevier}
}

@article{haltaufderheide2024ethics,
  title={The ethics of ChatGPT in medicine and healthcare: a systematic review on Large Language Models (LLMs)},
  author={Haltaufderheide, Joschka and Ranisch, Robert},
  journal={NPJ digital medicine},
  volume={7},
  number={1},
  pages={183},
  year={2024},
  publisher={Nature Publishing Group UK London}
}

@article{gallifant2025tripod,
  title={The TRIPOD-LLM reporting guideline for studies using large language models},
  author={Gallifant, Jack and Afshar, Majid and Ameen, Saleem and Aphinyanaphongs, Yindalon and Chen, Shan and Cacciamani, Giovanni and Demner-Fushman, Dina and Dligach, Dmitriy and Daneshjou, Roxana and Fernandes, Chrystinne and others},
  journal={Nature Medicine},
  pages={1--10},
  year={2025},
  publisher={Nature Publishing Group US New York}
}

@article{palmer2016defining,
  title={Defining major trauma using the 2008 Abbreviated Injury Scale},
  author={Palmer, Cameron S and Gabbe, Belinda J and Cameron, Peter A},
  journal={Injury},
  volume={47},
  number={1},
  pages={109--115},
  year={2016},
  publisher={Elsevier}
}

@article{shafi2009trauma,
  title={The trauma quality improvement program of the American College of Surgeons Committee on Trauma},
  author={Shafi, Shahid and Nathens, Avery B and Cryer, Gill H and Hemmila, Mark R and Pasquale, Michael D and Clark, David E and Neal, Melanie and Goble, Sandra and Meredith, Wayne J and Fildes, John J},
  journal={Journal of the American College of Surgeons},
  volume={209},
  number={4},
  pages={521--530e1},
  year={2009},
  publisher={LWW}
}

@article{boehm2022harnessing,
  title={Harnessing multimodal data integration to advance precision oncology},
  author={Boehm, Kevin M and Khosravi, Pegah and Vanguri, Rami and Gao, Jianjiong and Shah, Sohrab P},
  journal={Nature Reviews Cancer},
  volume={22},
  number={2},
  pages={114--126},
  year={2022},
  publisher={Nature Publishing Group UK London}
}

@article{chang2025continuous,
  title={Continuous multimodal data supply chain and expandable clinical decision support for oncology},
  author={Chang, Jee Suk and Kim, Hyunwook and Baek, Eun Sil and Choi, Jeong Eun and Lim, Joon Seok and Kim, Jin Sung and Shin, Sang Joon},
  journal={npj Digital Medicine},
  volume={8},
  number={1},
  pages={128},
  year={2025},
  publisher={Nature Publishing Group UK London}
}

@article{li2025towards,
  title={Towards a holistic framework for multimodal LLM in 3D brain CT radiology report generation},
  author={Li, Cheng-Yi and Chang, Kao-Jung and Yang, Cheng-Fu and Wu, Hsin-Yu and Chen, Wenting and Bansal, Hritik and Chen, Ling and Yang, Yi-Ping and Chen, Yu-Chun and Chen, Shih-Pin and others},
  journal={Nature Communications},
  volume={16},
  number={1},
  pages={2258},
  year={2025},
  publisher={Nature Publishing Group UK London}
}

@article{zhao2024aligning,
  title={Aligning knowledge concepts to whole slide images for precise histopathology image analysis},
  author={Zhao, Weiqin and Guo, Ziyu and Fan, Yinshuang and Jiang, Yuming and Yeung, Maximus CF and Yu, Lequan},
  journal={npj Digital Medicine},
  volume={7},
  number={1},
  pages={383},
  year={2024},
  publisher={Nature Publishing Group UK London}
}

@article{cui2024scgpt,
  title={scGPT: toward building a foundation model for single-cell multi-omics using generative AI},
  author={Cui, Haotian and Wang, Chloe and Maan, Hassaan and Pang, Kuan and Luo, Fengning and Duan, Nan and Wang, Bo},
  journal={Nature Methods},
  volume={21},
  number={8},
  pages={1470--1480},
  year={2024},
  publisher={Nature Publishing Group US New York}
}

@article{savova2010mayo,
  title={Mayo clinical Text Analysis and Knowledge Extraction System (cTAKES): architecture, component evaluation and applications},
  author={Savova, Guergana K and Masanz, James J and Ogren, Philip V and Zheng, Jiaping and Sohn, Sunghwan and Kipper-Schuler, Karin C and Chute, Christopher G},
  journal={Journal of the American Medical Informatics Association},
  volume={17},
  number={5},
  pages={507--513},
  year={2010},
  publisher={BMJ Group BMA House, Tavistock Square, London, WC1H 9JR}
}

@article{johnson2019mimic,
  title={MIMIC-CXR-JPG, a large publicly available database of labeled chest radiographs},
  author={Johnson, Alistair EW and Pollard, Tom J and Greenbaum, Nathaniel R and Lungren, Matthew P and Deng, Chih-ying and Peng, Yifan and Lu, Zhiyong and Mark, Roger G and Berkowitz, Seth J and Horng, Steven},
  journal={arXiv preprint arXiv:1901.07042},
  year={2019}
}

@article{johnson2023mimic,
  title={MIMIC-IV, a freely accessible electronic health record dataset},
  author={Johnson, Alistair EW and Bulgarelli, Lucas and Shen, Lu and Gayles, Alvin and Shammout, Ayad and Horng, Steven and Pollard, Tom J and Hao, Sicheng and Moody, Benjamin and Gow, Brian and others},
  journal={Scientific data},
  volume={10},
  number={1},
  pages={1},
  year={2023},
  publisher={Nature Publishing Group UK London}
}

@inproceedings{li2023blip,
  title={Blip-2: Bootstrapping language-image pre-training with frozen image encoders and large language models},
  author={Li, Junnan and Li, Dongxu and Savarese, Silvio and Hoi, Steven},
  booktitle={International conference on machine learning},
  pages={19730--19742},
  year={2023},
  organization={PMLR}
}

@article{alayrac2022flamingo,
  title={Flamingo: a visual language model for few-shot learning},
  author={Alayrac, Jean-Baptiste and Donahue, Jeff and Luc, Pauline and Miech, Antoine and Barr, Iain and Hasson, Yana and Lenc, Karel and Mensch, Arthur and Millican, Katherine and Reynolds, Malcolm and others},
  journal={Advances in neural information processing systems},
  volume={35},
  pages={23716--23736},
  year={2022}
}

@article{peng2023kosmos,
  title={Kosmos-2: Grounding multimodal large language models to the world},
  author={Peng, Zhiliang and Wang, Wenhui and Dong, Li and Hao, Yaru and Huang, Shaohan and Ma, Shuming and Wei, Furu},
  journal={arXiv preprint arXiv:2306.14824},
  year={2023}
}

@article{driess2023palm,
  title={Palm-e: An embodied multimodal language model},
  author={Driess, Danny and Xia, Fei and Sajjadi, Mehdi SM and Lynch, Corey and Chowdhery, Aakanksha and Wahid, Ayzaan and Tompson, Jonathan and Vuong, Quan and Yu, Tianhe and Huang, Wenlong and others},
  year={2023}
}

@inproceedings{girdhar2023imagebind,
  title={Imagebind: One embedding space to bind them all},
  author={Girdhar, Rohit and El-Nouby, Alaaeldin and Liu, Zhuang and Singh, Mannat and Alwala, Kalyan Vasudev and Joulin, Armand and Misra, Ishan},
  booktitle={Proceedings of the IEEE/CVF conference on computer vision and pattern recognition},
  pages={15180--15190},
  year={2023}
}

@inproceedings{han2024onellm,
  title={Onellm: One framework to align all modalities with language},
  author={Han, Jiaming and Gong, Kaixiong and Zhang, Yiyuan and Wang, Jiaqi and Zhang, Kaipeng and Lin, Dahua and Qiao, Yu and Gao, Peng and Yue, Xiangyu},
  booktitle={Proceedings of the IEEE/CVF Conference on Computer Vision and Pattern Recognition},
  pages={26584--26595},
  year={2024}
}

@article{liu2023visual,
  title={Visual instruction tuning},
  author={Liu, Haotian and Li, Chunyuan and Wu, Qingyang and Lee, Yong Jae},
  journal={Advances in neural information processing systems},
  volume={36},
  pages={34892--34916},
  year={2023}
}

@misc{jiang2023mistral7b,
      title={Mistral 7B}, 
      author={Albert Q. Jiang and Alexandre Sablayrolles and Arthur Mensch and Chris Bamford and Devendra Singh Chaplot and Diego de las Casas and Florian Bressand and Gianna Lengyel and Guillaume Lample and Lucile Saulnier and Lélio Renard Lavaud and Marie-Anne Lachaux and Pierre Stock and Teven Le Scao and Thibaut Lavril and Thomas Wang and Timothée Lacroix and William El Sayed},
      year={2023},
      eprint={2310.06825},
      archivePrefix={arXiv},
      primaryClass={cs.CL},
      url={https://arxiv.org/abs/2310.06825}, 
}

@article{xia2024mmed,
  title={Mmed-rag: Versatile multimodal rag system for medical vision language models},
  author={Xia, Peng and Zhu, Kangyu and Li, Haoran and Wang, Tianze and Shi, Weijia and Wang, Sheng and Zhang, Linjun and Zou, James and Yao, Huaxiu},
  journal={arXiv preprint arXiv:2410.13085},
  year={2024}
}

@inproceedings{xia2024rule,
  title={Rule: Reliable multimodal rag for factuality in medical vision language models},
  author={Xia, Peng and Zhu, Kangyu and Li, Haoran and Zhu, Hongtu and Li, Yun and Li, Gang and Zhang, Linjun and Yao, Huaxiu},
  booktitle={Proceedings of the 2024 Conference on Empirical Methods in Natural Language Processing},
  pages={1081--1093},
  year={2024}
}

@article{thawkar2023xraygpt,
  title={Xraygpt: Chest radiographs summarization using medical vision-language models},
  author={Thawkar, Omkar and Shaker, Abdelrahman and Mullappilly, Sahal Shaji and Cholakkal, Hisham and Anwer, Rao Muhammad and Khan, Salman and Laaksonen, Jorma and Khan, Fahad Shahbaz},
  journal={arXiv preprint arXiv:2306.07971},
  year={2023}
}

@article{guo2024prompting,
  title={Prompting Medical Large Vision-Language Models to Diagnose Pathologies by Visual Question Answering},
  author={Guo, Danfeng and Terzopoulos, Demetri},
  journal={arXiv preprint arXiv:2407.21368},
  year={2024}
}

@article{zhu2025guiding,
  title={Guiding Medical Vision-Language Models with Explicit Visual Prompts: Framework Design and Comprehensive Exploration of Prompt Variations},
  author={Zhu, Kangyu and Qin, Ziyuan and Yi, Huahui and Jiang, Zekun and Lao, Qicheng and Zhang, Shaoting and Li, Kang},
  journal={arXiv preprint arXiv:2501.02385},
  year={2025}
}

@article{yang2025medical,
  title={Medical Large Vision Language Models with Multi-Image Visual Ability},
  author={Yang, Xikai and Miao, Juzheng and Yuan, Yuchen and Wang, Jiaze and Dou, Qi and Li, Jinpeng and Heng, Pheng-Ann},
  journal={arXiv preprint arXiv:2505.19031},
  year={2025}
}

@misc{gao2024clinical,
  title={Clinical natural language processing for secondary uses},
  author={Gao, Yanjun and Mahajan, Diwakar and Uzuner, {\"O}zlem and Yetisgen, Meliha},
  journal={Journal of biomedical informatics},
  volume={150},
  pages={104596},
  year={2024},
  publisher={Elsevier}
}

@article{ouyang2019analysis,
  title={Analysis of the human protein atlas image classification competition},
  author={Ouyang, Wei and Winsnes, Casper F and Hjelmare, Martin and Cesnik, Anthony J and {\AA}kesson, Lovisa and Xu, Hao and Sullivan, Devin P and Dai, Shubin and Lan, Jun and Jinmo, Park and others},
  journal={Nature methods},
  volume={16},
  number={12},
  pages={1254--1261},
  year={2019},
  publisher={Nature Publishing Group US New York}
}

@article{bergquist2023evaluation,
  title={Evaluation of crowdsourced mortality prediction models as a framework for assessing artificial intelligence in medicine},
  author={Bergquist, Timothy and Schaffter, Thomas and Yan, Yao and Yu, Thomas and Prosser, Justin and Gao, Jifan and Chen, Guanhua and Charzewski, {\L}ukasz and Nawalany, Zofia and Brugere, Ivan and others},
  journal={Journal of the American Medical Informatics Association},
  volume={31},
  number={1},
  pages={35--44},
  year={2023},
  publisher={Oxford Academic}
}

@article{bergquist2023framework,
  title={A framework for future national pediatric pandemic respiratory disease severity triage: the HHS pediatric COVID-19 data challenge},
  author={Bergquist, Timothy and Wax, Marie and Bennett, Tellen D and Moffitt, Richard A and Gao, Jifan and Chen, Guanhua and Telenti, Amalio and Maher, M Cyrus and Bartha, Istvan and Walker, Lorne and others},
  journal={Journal of Clinical and Translational Science},
  volume={7},
  number={1},
  pages={e175},
  year={2023},
  publisher={Cambridge University Press}
}

@inproceedings{ben2019vqa,
  title={Vqa-med: Overview of the medical visual question answering task at imageclef 2019},
  author={Ben Abacha, Asma and Hasan, Sadid A and Datla, Vivek V and Demner-Fushman, Dina and M{\"u}ller, Henning},
  booktitle={Proceedings of CLEF (Conference and Labs of the Evaluation Forum) 2019 Working Notes},
  year={2019},
  organization={9-12 September 2019}
}

@article{huang2025survey,
  title={A survey on hallucination in large language models: Principles, taxonomy, challenges, and open questions},
  author={Huang, Lei and Yu, Weijiang and Ma, Weitao and Zhong, Weihong and Feng, Zhangyin and Wang, Haotian and Chen, Qianglong and Peng, Weihua and Feng, Xiaocheng and Qin, Bing and others},
  journal={ACM Transactions on Information Systems},
  volume={43},
  number={2},
  pages={1--55},
  year={2025},
  publisher={ACM New York, NY}
}

@inproceedings{nath2025vila,
  title={Vila-m3: Enhancing vision-language models with medical expert knowledge},
  author={Nath, Vishwesh and Li, Wenqi and Yang, Dong and Myronenko, Andriy and Zheng, Mingxin and Lu, Yao and Liu, Zhijian and Yin, Hongxu and Law, Yee Man and Tang, Yucheng and others},
  booktitle={Proceedings of the Computer Vision and Pattern Recognition Conference},
  pages={14788--14798},
  year={2025}
}

@article{he2024gsco,
  title={GSCo: Towards Generalizable AI in Medicine via Generalist-Specialist Collaboration},
  author={He, Sunan and Nie, Yuxiang and Wang, Hongmei and Yang, Shu and Wang, Yihui and Cai, Zhiyuan and Chen, Zhixuan and Xu, Yingxue and Luo, Luyang and Xiang, Huiling and others},
  journal={arXiv preprint arXiv:2404.15127},
  year={2024}
}

@article{caffagni2024revolution,
  title={The revolution of multimodal large language models: a survey},
  author={Caffagni, Davide and Cocchi, Federico and Barsellotti, Luca and Moratelli, Nicholas and Sarto, Sara and Baraldi, Lorenzo and Cornia, Marcella and Cucchiara, Rita},
  journal={arXiv preprint arXiv:2402.12451},
  year={2024}
}

@article{lupyan2016centrality,
  title={The centrality of language in human cognition},
  author={Lupyan, Gary},
  journal={Language Learning},
  volume={66},
  number={3},
  pages={516--553},
  year={2016},
  publisher={Wiley Online Library}
}
\end{refsection}

\end{document}